%% file: main.tex
\documentclass[a4paper, 11pt]{article}
\usepackage[english]{babel}
\usepackage[utf8]{inputenc}
\usepackage[T1]{fontenc}
\usepackage{csquotes}
\usepackage{array}
\usepackage{booktabs}
\usepackage{longtable}
\usepackage{cellspace}
\usepackage[theme=default-plain]{jlcode}

\usepackage{arxiv}

\usepackage{physics}
\usepackage{amsmath}
\usepackage[colorlinks=true, allcolors=blue]{hyperref}
\usepackage{amssymb}
\usepackage{mathtools}
\usepackage{ mathrsfs }
\usepackage{bbm}
\usepackage{bm}

\usepackage[backend=biber, 
            natbib=true,
            style=numeric-comp,
            sorting=none]{biblatex}

\renewcommand{\shorttitle}{A Systematic Exploration of Reservoir Computing}

\title{A Systematic Exploration of Reservoir Computing for Forecasting Complex Spatiotemporal Dynamics}

\author{Jason A. Platt \thanks{Department of Physics, University of California San Diego} \and Stephen G. Penny \thanks{Cooperative Institute for Research in Environmental Sciences,
University of Colorado Boulder,
Boulder, CO 80309,
Physical Sciences Laboratory,
National Oceanic and Atmospheric Administration,
Boulder, CO 80305}
\and Timothy A. Smith \footnotemark[2]
\and Tse-Chun Chen \footnotemark[2]
\and Henry D. I. Abarbanel \footnotemark[1]$^{\ }$ 
\thanks{Marine Physical Laboratory,
Scripps Institution of Oceanography,
University of California San Diego,
9500 Gilman Drive,
La Jolla, CA 92093, USA}}
\date{}
\input{defs}

\begin{document}

\maketitle
\begin{abstract}
A reservoir computer (RC) is a type of simplified recurrent neural network architecture that has demonstrated success in the prediction of spatiotemporally chaotic dynamical systems. A further advantage of RC is that it reproduces intrinsic dynamical quantities essential for its incorporation into numerical forecasting routines such as the ensemble Kalman filter---used in numerical weather prediction to compensate for sparse and noisy data. We explore here the architecture and design choices for a ``best in class'' RC for a number of characteristic dynamical systems, and then show the application of these choices in scaling up to larger models using localization. Our analysis points to the importance of large scale parameter optimization. We also note in particular the importance of including input bias in the RC design, which has a significant impact on the forecast skill of the trained RC model. In our tests, the the use of a nonlinear readout operator does not affect the forecast time or the stability of the forecast. The effects of the reservoir dimension, spinup time, amount of training data, normalization, noise, and the RC time step are also investigated. While we are not aware of a generally accepted best reported mean forecast time for different models in the literature, we report over a factor of 2 increase in the mean forecast time compared to the best performing RC model of \cite{Vlachas20} for the 40 dimensional spatiotemporally chaotic Lorenz 1996 dynamics, and we are able to accomplish this using a smaller reservoir size.
\end{abstract}

\section{Introduction}
Chaotic nonlinear dynamical systems are deterministic, but have the property that even small uncertainties in initial conditions lead to exponentially growing errors that limit any forecast of such a system. A typical linear systems analysis (\eg, a fourier transform) of such signals would conclude that the signal is noise, thus missing an essential feature of chaos. An illustrative example of this is the error made by Edwin Colpitts~\cite{colpitts18} in 1918---who designed and built a nonlinear variable frequency circuit (then made with vacuum tubes, now using bipolar transistors)---in identifying the irregularity in the output of his circuit as ``noise.'' 76 years later Michael Kennedy~\cite{Kennedy94} recognized this feature to be deterministic chaos. 

The original identification of chaotic motion of nonlinear systems was made by Poincare in the 19th century while analyzing the three body problem of classical mechanics. The present era of understanding and insight into the properties of such systems was initiated by Edward Lorenz in 1963~\cite{Lorenz63}. Lorenz' simplification of convection in the Earth's lower atmosphere introduced the idea of deterministic, nonperiodic behavior.

The study of nonlinear dynamical systems typically requires the use of numerical integration methods, as analytic solutions to nonlinear differential equations are not typically available. This is an important contrast to the well established analysis of linear dynamics. The solution to the linear filtering problem with discrete data proposed by Kalman in 1960\cite{Kalman60}---now called the Kalman filter---is still the basis for many of the most advanced algorithms for predicting chaotic systems, especially in the highly nonlinear application of numerical weather prediction (NWP)~\cite{even09}.

We do note here that there are also fully nonlinear developments of data assimilation (in contrast with the linearity assumptions in the Kalman filter) in working with the dynamics of the Earth system for weather and climatic studies~\cite{abar2021}. The use of these techniques in NWP faces many challenges. However, in other arenas such as the behavior of single biophysical neurons, where the dynamics are many orders of magnitude smaller than NWP, the challenges are smaller and the effectiveness of the tools much greater \cite{Miller18}.

One limitation of the linear Kalman filter and its nonlinear descendants is that the technique requires a detailed physical model of the underlying process---something that is usually not known for complicated physical systems. To address this issue, there has been an increasing drive towards replacing the physical model altogether by using a ``data-driven model'' as the predictor. Such data-driven models fall into the category of machine learning (ML); examples include autoregressive methods (ARMA), the multilayer perceptron (MLP), and the recurrent neural network (RNN) \cite{Goodfellow16}.

RNNs are a family of artificial neural networks where the internal state of the network depends explicitly both on the previous internal or `hidden' state and an external driving signal. Thus, there is a natural ordering to the data---we call the order label time---which explicitly allows treatment of an RNN as a dynamical system. In contrast, other forms of artificial neural networks impose no ordering on the data and thus their natural mathematical description is as a function \cite{Lipton15}.

An RNN takes a $D$-dimensional input sequence $\u(t) \in \mathbb{R}^D$, evolves as the $N$-dimensional state $\r(t)\in \mathbb{R}^N$, and is asked to predict a value $\vb y(t) \in \mathbb{R}^Y$. Equation 10.5 in \textit{Deep Learning} by Goodfellow, et.al. \cite{Goodfellow16}, gives the general discrete time RNN evolution equation
\begin{equation}
    \r(t) = F(\r(t-1), \u(t-1); \theta) \label{eq: RNN}
\end{equation}
where $\theta$ represents the RNN hyperparameters. This shows the dependence of $\r(t)$ on the previous RNN state $\r(t-1)$ and external driving signal $\u(t-1)$. The output/readout layer of the network then `reads' $\r(t)$ as containing the information on previous inputs, through the recursive relation, to predict $\vb y(t)$. In other words, the output of the network $\hat{\u}(t) \equiv \Wout (\r(t))$ is a function $\Wout$ of the RNN hidden state $\r(t)$. The operator $\Wout$ defined here is a generic function, and could even be implemented as a neural network. Later we will take $\Wout$ to be a linear operator (a matrix). We can thus use $\hat \u(t) \sim \vb y(t) = \u(t)$ as a forecast.

Long short-term memory (LSTM) networks, introduced by Hochreiter and Schmidhuber (1997) \cite{Hochreiter97} and Gers et al. (1999) \cite{Gers2000} have been one of the most useful varieties of RNN, as they are capable of maintaining a representation of long-term dependencies. This design helped LSTMs to circumvent a number of difficulties that plagued RNNs regarding exploding and vanishing gradients during optimization using backpropagation. A simplified form of the LSTM was later developed by Kyunghyun Cho et al. in 2014 \cite{Kyunghyun14} called the Gated Recurrent Unit (GRU), which has a reduced set of trainable model parameters.

\textbf{Our purpose here is to systematically explore an alternative form of RNN called a reservoir computer (RC) due to its demonstrated success in the prediction of chaotic dynamics. Through the examination of architectural and procedural choices for the RC design, we present ``best in class'' results for a wide range of simple dynamical systems. We then explore the method of localization for scaling the RC to larger spatiotemporally chaotic systems.} We note that because the structure of the RC is nearly identical to the canonical RNN \cite{elman1990}, many of the findings will be relevant for general RNNs as well. 

\subsection{Background}
RC \cite{luk09, Jaeger12} is a general term that encompasses multiple research tracks that developed in different fields in the early 2000's, in particular Echo State Networks (ESNs) \cite{Jaeger01, Jaeger02, Jaeger04} and Liquid State Machines (LSMs) \cite{Maass02}. It is a kind of RNN with a structure consisting of a fixed high-dimensional ``reservoir'' combined with a single trained output layer. The fixed nature of the reservoir reduces the size of the trained parameter space to a handful of `macro-scale' parameters governing global properties, as well as `micro-scale' parameters that comprise the matrix and vector elements. A key advantage of RC is that the final output or `readout' layer can be formulated as a linear layer and then trained using linear regression. This reduction in the size of the searchable parameter space allows one to easily bypass the exploding/vanishing gradient problems that arise due to the use of backpropagation schemes \cite{Hochreiter01, Bengio93, Pascanu13}.

In 2007, Schrauwen, Verstraeten, and Campenhout \cite{Schrauwen07} drew the connection between methods proposed by \cite{Jaeger01} and \cite{Maass02} and the state-of-the-art learning rule for RNNs described by \cite{Steil2004}. For example, the recurrent learning rule of Atiya and Parlos (2000) \cite{Atiya2000} trains the output weights while the internal weights are only globally scaled up or down.

In addition to being easy to use and train, RCs have been shown to be very successful for chaotic time series prediction \cite{pathak18, Lu18, Griffith19, Vlachas20}. For NWP specifically, a major use case of multivariate chaotic time series prediction, Arcomano et.al. (2020) \cite{arcomano20} applied RC to a global atmospheric forecast model. Penny et.al. (2021) \cite{penny2021} integrated RC with state-of-the-art data assimilation methods to estimate the forecast error covariance matrix, tangent linear model representation, and other components of the data assimilation process that are typically costly to either formulate or compute. Further computational advantage are afforded by implementing RCs on GPUs or even dedicated hardware \cite{canaday18}.





We note that other ML methods have been explored for modeling dynamical systems, with varying degrees of success. While we cannot name them all, we highlight a few examples. Bocquet et al. \cite{bocquet2020, bocquet2021} used NNs and Convolutional NNs \cite{Goodfellow16} to learn ODEs that could then be integrated using traditional numerical methods. As with RC, they have similarly been able to reconstruct the Lyapunov spectrum of simple systems. Convolutional LSTMs have been used in MetNet \cite{sønderby2020metnet} to predict the evolution of precipitation. Gauthier et.al. \cite{gauthier21} introduced a vector autoregression scheme that has shown remarkable success in the prediction of small systems; see, for example, Clark et.al.  \cite{Clark21} for an application in neurobiology.  Data assimilation has been shown equivalent to feedforward Machine learning~\cite{abar18}.

\subsection{Why a Systematic Exploration of RC is Needed}
Although published guides to the general use of RC exist \cite{Lukoševičius12} and there have been multiple studies on specific applications of RC with dynamical systems \cite{Tanaka19}, there has been little guidance on best practices for implementing RC for forecasting chaotic dynamical systems. A practitioner must piece together design decisions by drawing on potentially conflicting information from multiple sources. We intend here to point the researcher or practitioner to strategies that have proven more successful, and to avoid those that we show to be `traps' (\ie, commonly published design decisions that have little or even negative impact on overall performance). Additionally, using ML to forecast chaotic dynamical systems presents a number of its own unique challenges when compared to standard practices in ML. There is a high potential for misunderstanding by practitioners and developers.

This paper aims to lay out in clear terms the problem being solved by RC for the specific task of forecasting chaotic dynamical systems, guidance on how to train and test the RC models, as well as the lessons we have learned in applying these methods over the last several years on the specifics of implementations for particular problems.

\subsection{Structure}
We proceed as follows:
\begin{enumerate}
    \item Section \ref{sec: def} gives an overview of the RC implementation.
    \item Section \ref{sec: theory} explains how an RC works and why it has properties, such as the reproduction of error growth statistics, that may not be well documented in other ML models.
    \item Section \ref{sec: testing} gives details of the generative models and advice on how to properly test a new RC implementation.
    \item Section \ref{sec: training} details the general training procedure and shows experimental results on experimental design choices.
    \item Section \ref{sec: scaling up} discusses scaling up the RC to high dimensional systems.
\end{enumerate}

\section{RC Definition} \label{sec: def}
In the context of chaotic time series prediction we apply RC to the temporal ML problem where the task is to predict a time series $\u(t)$ generated from an autonomous dynamical system 
\begin{equation}
    \dot \u(t) = f_u(\u(t)),
\end{equation}
where the dot denotes a time derivative and $f_u$ denotes the equations of the dynamical system. The dimension of the input system is $D$. In general $f_u$ is not known, nor does it need to be known for RC to be used. We do assume that $f_u$ exists and that time dependent data can therefore be generated deterministically. The system $f_u$ can describe any physical process and therefore these techniques can be applied in fields such as biology, hydrology, meteorology, oceanography, economics, chemistry, and many others.

Here we use an RC model with the specific form of a simplified ``Elman'' style\cite{elman90} RNN. The RC consists of three layers: an input layer $\Win$, the reservoir itself, and an output layer $\Wout$. The reservoir is composed of $N$ nodes that are generally acted upon with simple nonlinear elements \eg, $\tanh$ activation functions. The nodes in the network are connected through an $N \times N$ adjacency matrix $\A$, chosen randomly to have a connection density $\rho_A$ and non-zero elements uniformly chosen between $[-1, 1]$, scaled such that the maximal eigenvalue of $\A$ is a number denoted the spectral radius ($\rho_{SR}$).

The input layer $\Win$ is an $N \times D$ dimensional matrix that maps the input signal $\u(t)$ from $D$ dimensions into the $N$ dimensional reservoir space. The elements of $\Win$ are chosen uniformly between $[-\sigma, \sigma]$. The matrix $\Wout$ provides the mapping from the RC reservoir state to the system state as $\hat \u(t) = \Wout \r(t)$. The reservoir state $\r(t)$ can be viewed as embedding the information given in the time series $\u(t), \u(t-1),\ldots,\u(0)$ in a higher $N > D$ dimensional space, consistent with Takens theorem of time-delay embedding \cite{Hart20}. Takens theorem implies that information from unobserved states of the dynamics is contained in the time delay signal, thus allowing an RC to operate even when not all the information is measured.

\begin{figure}[!htpb]
    \centering
    \includegraphics[width = 0.85\textwidth]{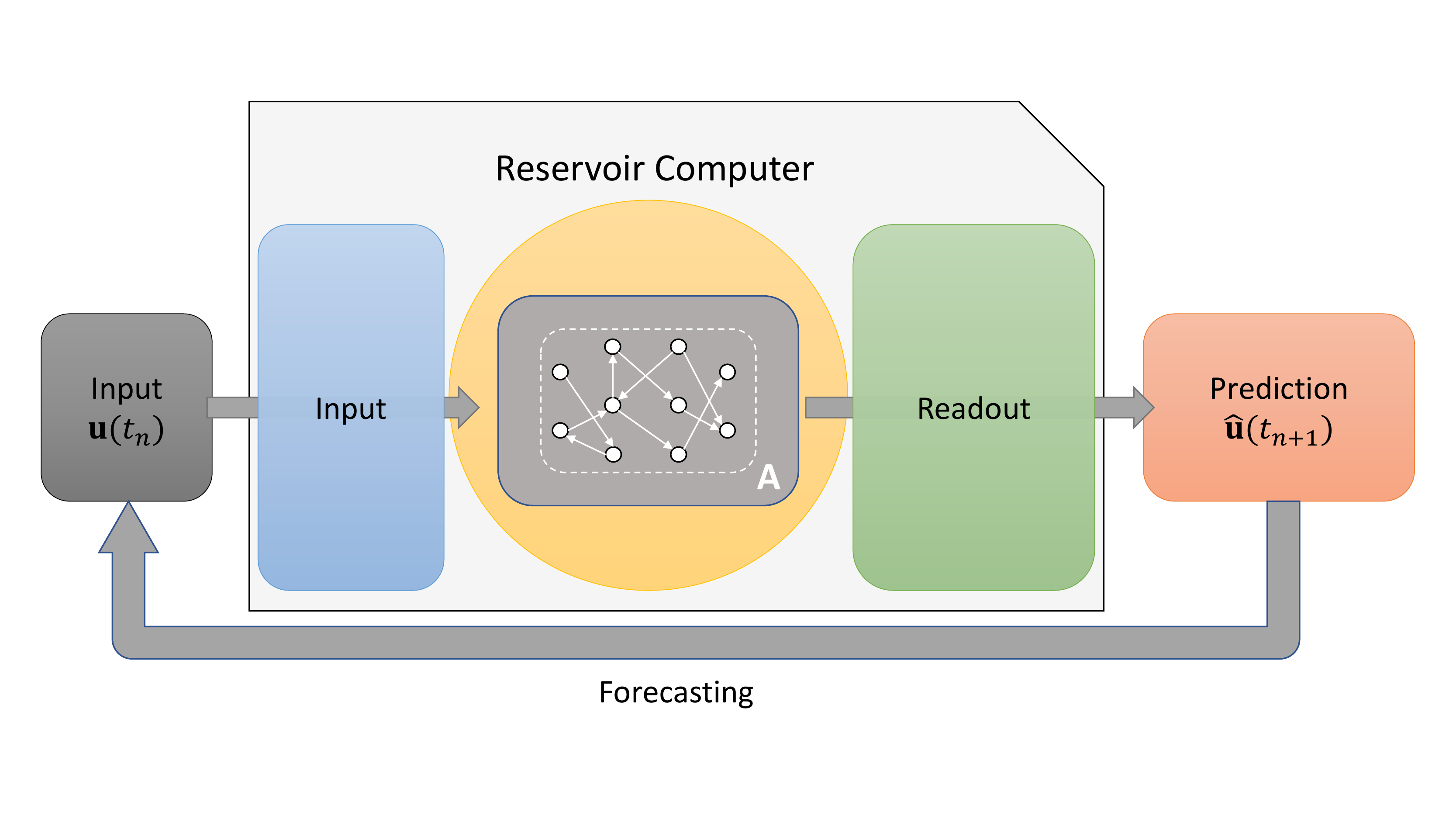}
    \caption{Schematic of information flow in the RC.  The input data $\u(t_n) \in \mathbb{R}^D$ is mapped through the input layer $\Win \in \mathbb{R}^{N \times D}$ into the reservoir $\r \in \mathbb{R}^N$ where the nonlinear activation function (e.g. $tanh$) is applied and mixed through a fixed (i.e. not trained) adjacency matrix $\A \in \mathbb{R}^{N \times N}$. The readout layer $\Wout \in \mathbb{R}^{D \times N}$ is typically trained using linear regression and gives the one-step-ahead prediction. Longer-term forecasts are achieved by cycling a feedback of the output back to the input.}
    \label{fig: reservoir_def}
\end{figure}

\subsubsection{Training $\Wout$}

The output layer $\Wout \in \mathbb{R}^{D \times N}$ is a matrix such that $\Wout \r(t) \equiv \hat \u(t) \sim \u(t)$ chosen during the training phase. The elements of $\Wout$ are what we call ``micro-scale'' parameters---in contrast to the ``macro-scale'' parameters ($\sigma$, $\rho_{SR}$, $\rho_A$, N,\ldots) that set overall properties of the RC by adjusting multiple matrix elements at once. An analogy would be that setting the macro-scale parameters is similar to specifying the overall temperature or pressure of a gas, the positions and velocities of each individual particle need not be taken into account, greatly simplifying the problem. The micro-scale parameters that comprise the elements of $\mathbf{W}_{out}$ can be trained at minimal cost using linear regression.


The reservoir dynamics themselves can either be posed in mapping or differential form. The differential form is 
\begin{equation}
    \dot \r(t) = f_{r}^d(\r(t), \u(t)) = \gamma \mqty[-\r(t) +\tanh(\A \r(t) + \Win \u(t)+ \sigma_b \mathbbm{1})] \label{eq: differential_form}
\end{equation} with $\gamma$ the inverse time constant, $\sigma_b$ the input bias, $\mathbbm{1}$ a vector of 1's, and $t \in \mathbb{R}$.

The mapping form is
\begin{equation}
    \r(t+1) = F_r^d(\r(t), \u(t)) = \alpha \tanh(\A \r(t) + \Win \u(t)+ \sigma_b \mathbbm{1}) + (1 - \alpha) \r(t) \label{eq: mapping_form}
\end{equation}
with $t \in \mathbb{Z}$. The notation is as follows; $f$ describes the differential form, while $F$ describes the mapping form. $F^d_r$ denotes the `driven' RC, as opposed to `autonomous' ($F^a_r$), $\r$ is the reservoir state vector, and  $\u$ is the state produced by the true dynamics, which is provided as input data.  Eqs.(\ref{eq: differential_form}, \ref{eq: mapping_form}) are denoted the ``driven RC'' because $\u(t)$ enters directly into the reservoir equation of motion.

The mapping form is equivalent to the differential form when the latter is integrated using the Euler method. This can be shown by setting $\alpha = \gamma \Delta t$ and discretizing Eq.(\ref{eq: differential_form}) with the Euler discretization scheme. While both the mapping and differential forms can be used interchangeably, this detail must be acknowledged as the integration method may affect the RC accuracy. Alternative forms of the RC have been suggested---see \cite{Tanaka19} for a review---but so far we have not found any advantages in using those forms.  

We determine the micro-scale parameters in $\Wout$ by optimizing the loss
\begin{equation}
\begin{array}{ll}
\mbox{minimize} & \mathscr{L}_{\rm micro}(\r, \u) = \mbox{$||\Wout \r - \u||^2 + \beta ||\Wout||^2$,} \\
\end{array} \label{eq: loss_micro}
\end{equation}
where $\r$ is a matrix containing $\r(t)$ and $\u$ is a matrix containing $\u(t)$ for all $t$ in the training dataset, and $\beta$ is a Tikhonov-Miller regularization hyperparameter \cite{tikhonov77}. This is also known as ridge regression. The solution is
\begin{equation}
    \Wout = \u \r^T(\r\r^T + \beta \mathbb{I})^{-1}, \label{eq: sol_linear_reg}
\end{equation}
where $\I$ is the $N \times N$ identity matrix. There is an option to add nonlinearity to the readout operator, which is addressed in section \ref{sec: readout}.


\subsubsection{Forecasting}
When making a forecast, the external forcing $\u(t)$ is replaced by a feedback loop $\Wout \r(t)$. Thus the equation for the ``autonomous RC'' is
\begin{equation}
    \r(t+1) = F_r^a(\r(t)) = \alpha \tanh[\A \r(t) + \Win \Wout \r(t)+ \sigma_b \mathbbm{1}] + (1 - \alpha) \r(t). \label{eq: predicting_res}
\end{equation}
One can see that $\u(t)$ is replaced by $\Wout \r(t)$ and so in prediction we have $F_r^a(\r(t))$ being an autonomous dynamical system, allowing multi-timestep forecasts via recursion.

 \vspace{0.2in}
\begin{table}[htpb!]
\centering
\begin{tabular}{|c c|} 
 \hline

 {\bf Parameter} & {\bf Description} \\
 \hline\hline
$\rho_{SR}$ & Spectral radius (max eigenvalue) of the adjacency matrix $\A$ \\ 
  \hline
 $\rho_A$ & Density of the adjacency matrix $\A$ \\
  \hline
 N & Degrees-of-Freedom of the reservoir\\
 \hline
 $\alpha$ & Leak rate (time constant) of the reservoir computer \\
 \hline
 $\sigma$ & Strength of input signal \\
 \hline
  $\sigma_b$ & Strength of input bias \\ 
 \hline
 $\beta$ & Tikhonov regularization parameter \\
 \hline
\end{tabular}
\caption{Global scalar parameters of the RC. These parameters may either be treated as `macro-scale' model parameters to be optimized during training, or may be used as hyperparameters and tuned manually.} 
\label{tab: hyperparams}
\end{table}
\section{RC Theory} \label{sec: theory}

\subsection{Stability} \label{sec: stability}

Recalling the earliest works on ESNs \cite{Jaeger01, Jaeger02}, setting the spectral radius $\rho_{SR} < 1$ is given as a guarantee of the ``echo state property'' \ie, generalized synchronization and fading memory---see section \ref{sec: GS}. Over the years the caveats in the original derivation, for example that the condition holds only when $\u(t)=0$ is an input, have generally been lost; the claim is therefore that the $\rho_{SR}$ should always be set to a value less than 1. Despite this statement being proved empirically incorrect \cite{Verstraeten06, Jiang19}, it is usually one of the first statements one comes across when studying the RC literature. In the following we give an alternative analysis based on dynamical stability theory.


Intuitively, local stability around a fixed point requires that small perturbations do not result in large movements to the state of the system \cite{strogatz00}. If the RC were unstable, a small variation in training data would result in vastly different reservoir states, making finding a readout $\Wout$ practically if not theoretically impossible \cite{Verstraeten09}. Therefore finding conditions for local instability can in practice give us conditions on the trainability of the RC.

The fixed points of a nonlinear map $\r(n+1) = F_r(\r(n))$ are solutions of the equation $\r_\star = F_r(\r_\star)$. Therefore, for our reservoir Eq.(\ref{eq: mapping_form}) we find can find the equilibrium 
\begin{equation}
    \r_\star = \tanh(\A \r_\star + \Win \u + \sigma_b \mathbbm{1}), \label{eq: fixed_point}
\end{equation}
which has solution $\r_\star = \vb{0}$ when $\u = \vb{0}$, and $\sigma_b = 0$.

The stability of the system is governed by the eigenvalues of the Jacobian of the map evaluated at $\r_\star$. If the magnitude of the largest eigenvalue is $< 1$ then the system is stable around the fixed point \cite{Chen05}; one caveat is that local stability does not imply global stability where a perturbed orbit will stay in the neighborhood of the unperturbed orbit. The Jacobian of a nonlinear map is $\DF_r = \pdv{F_r}{r}\big |_{r_\star}$. For the RC, taking the derivative of Eq.(\ref{eq: mapping_form}) with respect to $\r$
\begin{equation*}
    \DF_r^d = \pdv{F_r}{r}\Big |_{r_\star} = \alpha \cdot \diag(\mathbbm{1} - \tanh(\A \r_\star + \Win \u+ \sigma_b \mathbbm{1})^2)\A + (1 - \alpha) \mathbb{I}
\end{equation*}
and plugging in $\r_\star = \tanh(\A \r_\star + \Win \u + \sigma_b \I)$ from Eq.\eqref{eq: fixed_point} to simplify, we find
\begin{equation}
    \DF_r^d = \pdv{F_r}{r}\Big |_{r_\star} = \alpha \cdot \diag(\mathbbm{1} - \r_\star^2)\A + (1 - \alpha) \mathbb{I}, \label{eq: driven Jacobian}
\end{equation}
where $\diag(\cdot)$ denotes the formation of a diagonal matrix.
\begin{figure}[!htpb]
    \centering
    \includegraphics[width=0.5\textwidth]{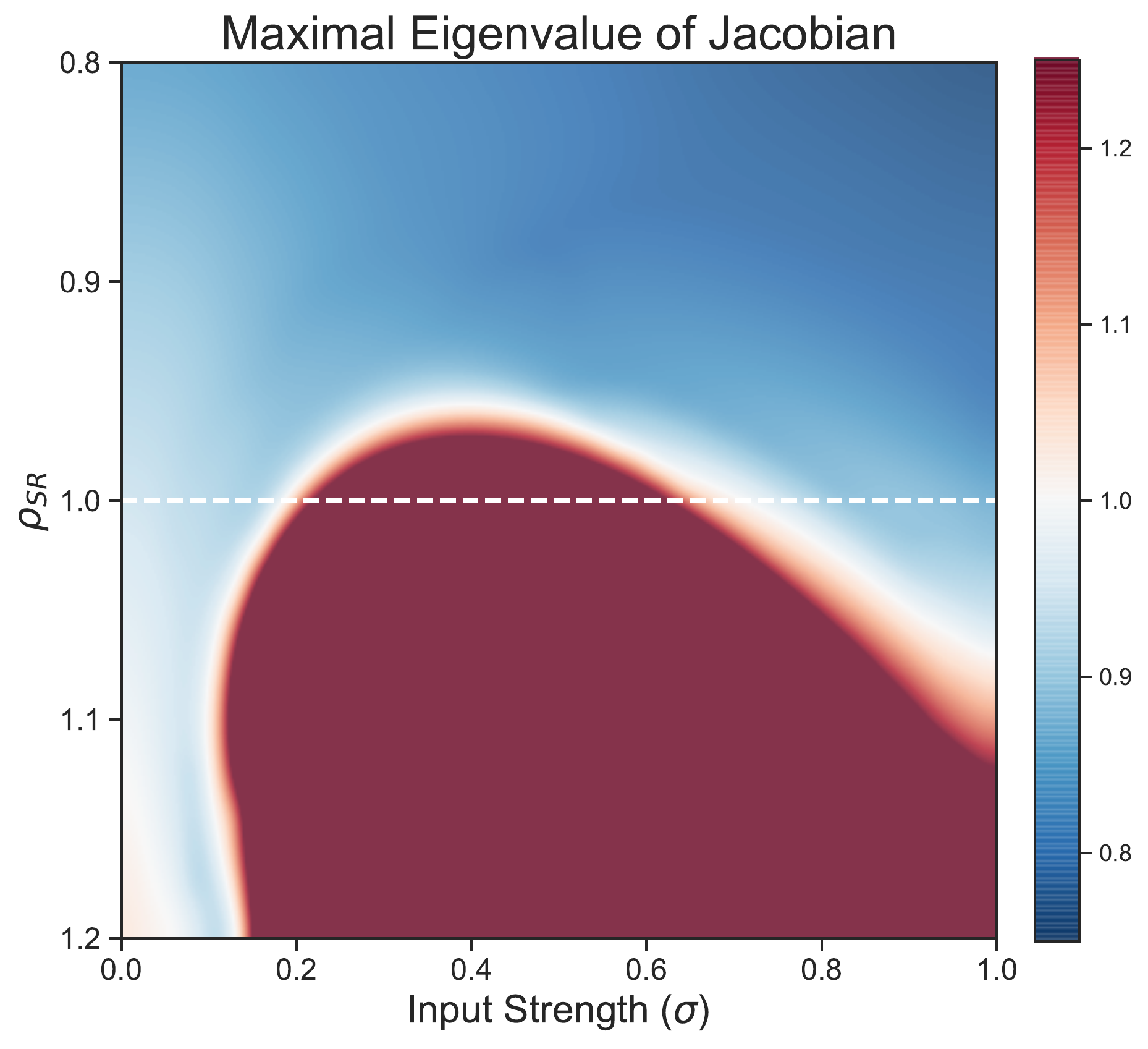}
    \caption{The magnitude of the maximal eigenvalue of the Jacobian matrix Eq.(\ref{eq: driven Jacobian})---$\mqty{N=200, \alpha = 0.5, \rho_A=0.9}$. When the eigenvalue of the Jacobian is $< 1$ then the fixed point equilibrium of the RC is stable; this is a general property of discrete dynamical systems \cite{Chen05}.  If the fixed point is stable then orbits of the RC as it is driven around the fixed point will tend to stay in the neighborhood of that point. This would suggest that the RC is trainable (\ie, a ``good'' $\Wout$ can be found)---see section \ref{sec: GS} for a more general criterion for trainability. An eigenvalue above 1 indicates an unstable fixed point, making the RC sensitive to perturbations (\eg, noise) in the driving signal.  The white line indicates $\rho_{SR} = 1$, the generally accepted limit for the spectral radius. This example indicates that one should be careful when assuming a requirement that $\rho_{SR} < 1$.}
    \label{fig: stability}
\end{figure}
We can use this equation to visualize the areas of stability of the RC. If we assume $\r^\star \sim 0$ for $\tilde \u = \Win \u$ small then $\tanh(A \r^\star + \tilde \u) \sim \tanh{\tilde \u}+\r^\star \diag(\sech^2(\tilde \u)) A$ and we solve for 
\begin{equation}
    \r^\star = \tanh(\tilde \u) (\mathbb{I} + \diag(\sech^2(\tilde \u))A)^{-1}. \label{eq: fixed_point_approx}
\end{equation}  With the fixed points given we can plot the maximal eigenvalue of $\DF_r^d$. An example is given in Fig.(\ref{fig: stability}). In this example, the RC is stable for values of $\rho_{SR}$ that are much greater than 1. This also gives a clear example that for nonzero input, there can be regions where $\rho_{SR}$ is less than one but the RC is unstable.

Having a globally stable RC is most likely necessary, but certainly not a sufficient condition for obtaining a $\Wout$ that gives good predictions.  For instance, it has been noted many times that the RC operates best at the ``edge of stability''\cite{Verstraeten09, Boedecker12, Carroll20} and in general the RC is sensitive to parameter variation. An unstable RC is, however, rather unlikely to have a linear map between $\r$ and $\u$.

\subsection{Attractor Reconstruction and Lyapunov Exponents}
The Lyapunov spectrum, composed of a system's Lyapunov Exponents (LE), is a tool to probe the fundamental characteristics of a dynamical system \cite{Lyapunov, Abarbanel96_book}. For example, one definition of a chaotic system is that there is at least one positive LE \cite{Eckmann85}. LEs can also be used to estimate the rate of entropy production (Kolmogorov-Sinai Entropy) and the fractal or information dimension \cite{Eckmann85}.

The LEs describe how two points $x_1$ and $x_2$, separated initially by a distance $\delta x_0 = |x_1 - x_2|$, evolve in time with respect to one another. More specifically,
\begin{equation*}
    \delta x(t) \approx \delta x_0 e^{\lambda_1 t}
\end{equation*}
with $\lambda_1$ being the largest LE \footnote{it is important to note that this is only actually true when $\delta x$ lies exactly along the direction $\lambda_1$, however for large $t$ the largest exponent will dominate.}. Therefore if $\lambda_1$ is positive, over time the two initial conditions that at first only differ by a small amount will diverge away from one another exponentially fast in the linearized regime---the definition of chaos. We give a more rigorous definition of LEs in the appendix along with a discussion of how to calculate them---Appendix \ref{app: LE}.

\subsubsection{Application to Reservoir Computing}
When the LEs of the \textbf{autonomous} Eq.\eqref{eq: predicting_res} RC match the LEs of the input data generated from the dynamical system $f_u$ then the RC is said to have ``reconstructed'' the attractor \cite{Lu18}. A reconstructed attractor is strongly correlated with properties such as the physicality and robustness of forecasts \cite{platt21}. The LEs of $f_u$ can be estimated from the data with no knowledge of the dynamics \cite{Abarbanel96_book}. 

To calculate the LEs of the autonomous reservoir we note that the Jacobian/linear propagator of Eq.\eqref{eq: predicting_res} is
\begin{equation}
    \DF_r^a(n) = \alpha \cdot \diag(\mathbbm{1}-\tanh[\W \r(n) + \sigma_b \mathbbm{1}]^2)\W + (1-\alpha)\mathbb{I}
\end{equation}
with diag denoting a diagonal matrix and the constant matrix $\W = \A +\Win \Wout$. This propagator may be used in the algorithms given in Geist et.al., \cite{Geist90} using the procedure given in \cite{Lu18, Vlachas20} to calculate the LEs.

\subsection{Generalized Synchronization} \label{sec: GS}
As described in detail by \cite{Lu18, platt21}, the RC works by making use of the concept of generalized synchronization (GS) \cite{sushchik95}. The definition of GS is: for a drive system $\u \in \mathbb{R}^D$ and response system $\r \in \mathbb{R}^N$, they are synchronized if there exists a function $\psi$ such that $\r = \psi(\u)$ \cite{sushchik95}. Intuitively this implies that the dynamics of the response system are entirely predictable from the history of the input; in RC, for a contracting system, this property is related to the ``echo state property'' \cite{Jaeger01} and ``fading memory'' \cite{boyd85}. The requirement to find a matrix $\Wout$ such that $\Wout \r(t) = \u(t)$ is tantamount to finding a linear approximation to the local inverse of the generalized synchronization function---we call this statement predictive generalized synchronization \cite{platt21}. We would like to note that the statement of the existence of the inverse of $\psi$ is in general not a global, but a local property due to the necessity of $\psi$ being differentiable \cite{Hunt97} for $\psi^{-1}$ to exist.  Grigoryeva et. al., \cite{Grigoryeva21a, Grigoryeva21b} details general conditions for differential GS to exist in an RC. When $\psi$ does not meet these criteria then the RC is untrainable; this statement is connected to the stability criterion explored in section \ref{sec: stability}.

The concept of GS is useful for visualizing the operation of the RC. When $\u(t)$ and $\r(t)$ are synchronized, the combined state $\mathbb{R}^{N+D}$ will lie on an invariant {\it synchronization manifold} $\M$~\cite{Pecora97}. This manifold is generally low dimensional \cite{platt21} and so the dynamics of the RC are taking place on this low dimension structure in phase space. In addition, the concept of synchronization shows that there is a certain amount of ``spinup'' time needed for the transient initial conditions to contract to the synchronization manifold.  The amount of time needed for spinup is directly related to the strength of the conditional Lyapunov exponents (CLEs) \cite{Pecora90} of the synchronization manifold---see appendix \ref{app: LE} for CLEs. The statement of stability of the motion on the manifold \cite{Pecora97, Pecora00} is that the contraction normal to the manifold must be larger than the contraction tangential to it \cite{Kocarev00}, thus giving a statement of the stability of the RC.

The CLEs and the constraint that they be negative is in fact closely related to the stability conditions we discuss in section \ref{sec: stability}. If the RC is unstable everywhere, $\DF_r^d > 1$, then the CLEs will always be positive and the RC will not synchronize with the input data. As a last note, it has been observed that RCs work best when at the edge of stability \cite{Verstraeten09, Boedecker12, Carroll20}; this edge corresponds to the largest CLE being $\sim 0$ and $\DF_r^d \sim 1$ \cite{platt21}.

\section{Data Generation for Numerical Experiments} \label{sec: testing}
We begin with a discussion of the training and testing data used in the following numerical experiments. The use of input data is different for dynamical systems forecasting than for the typical ML use case. Typically in ML there are standard training and testing datasets \eg, MNIST \cite{MNIST}, that one can acquire in order to directly test newly proposed models against published results from previous models \cite{ML_data_list}. For dynamical systems forecasting we instead have certain standard models from which one can generate data. Therefore it is important to make sure one is observing best practices when constructing ones own training, validation and testing sets.

\begin{figure}[!htpb]
    \centering
    \includegraphics[width=0.7\textwidth]{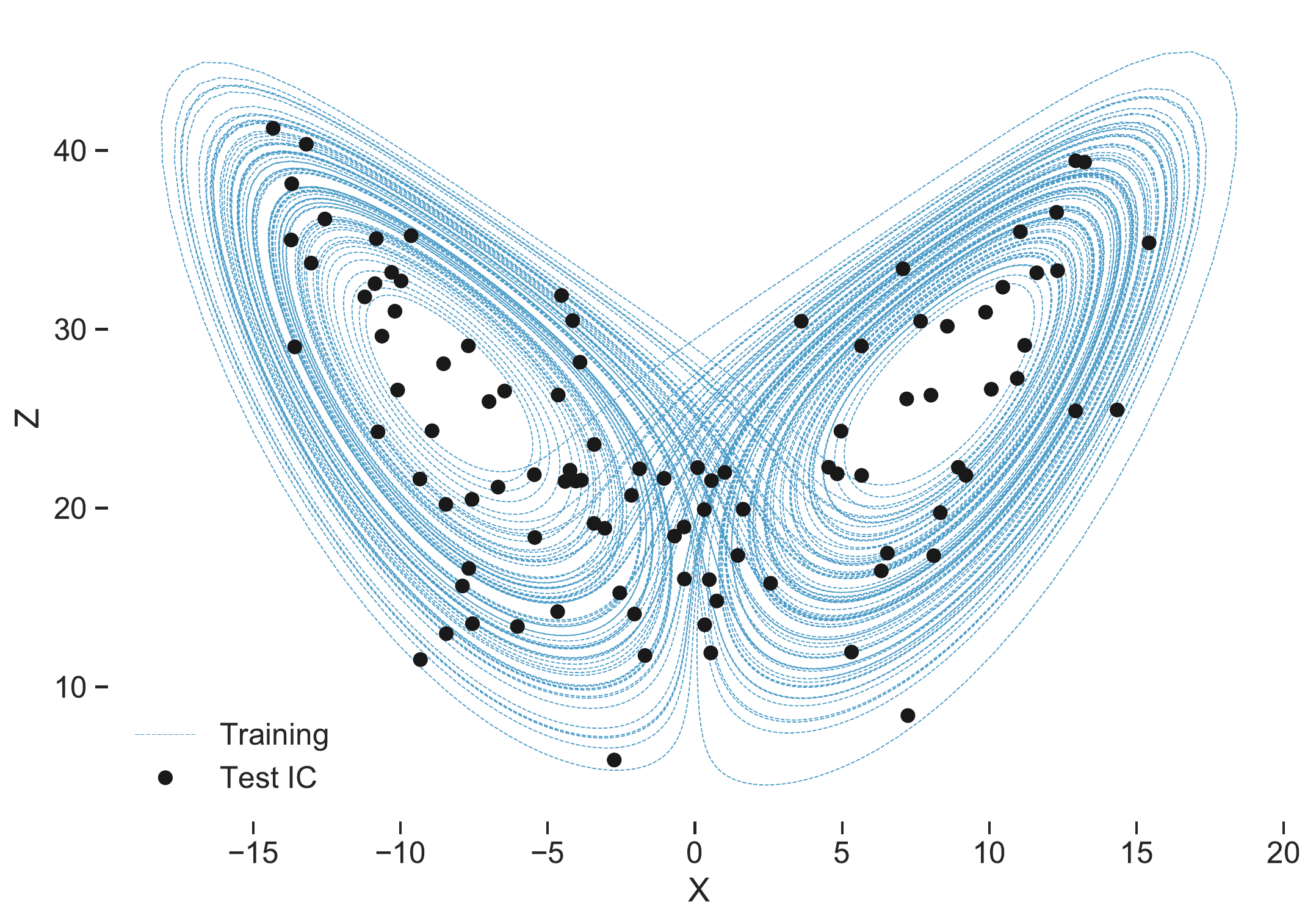}
    \caption{Training data (blue) and test forecast initial conditions (black) for the L63 system. For the training data it is important that the data cover most parts of the input attractor. This ensures that the RC does not overfit a local section of the dynamics and that it can generalize. The difference in dynamics over the attractor can be formalized by the finite time Lyapunov exponents (FTLE) \cite{Abarbanel92, Abarbanel96_book}, which denote the localized error growth rates. These can be positive or negative for the L63 system depending on the initial condition. Likewise for the testing data, the initial conditions for the test forecasts should widely sample the attractor. It is important that the testing data be long term forecasts and not one-step forecasts in order to capture the broadband frequencies of the chaotic dynamics.}
    \label{fig: L63_train_test}
\end{figure}

We would like to construct independent training, validation and testing data for training and testing our ML models. In order to be confident in our results, the validation/testing data must be independent from both one another and the training data. Furthermore, we would like the testing data to sample as much of the phase space of $f_u$ as possible, so that we can be confident that the RC generalizes to parts of the input with different properties than which it has been trained on. All data have been generated with these principles in mind Fig.(\ref{fig: L63_train_test}).

Additionally, we report forecast times in terms of the Lyapunov timescale of the input dynamical system. This timescale $\tau_\lambda = 1/\lambda_1$, gives the natural timescale of error growth in the system and is therefore a useful measure of how ``good'' the forecast is. Not even a perfect model can predict a chaotic dynamical system forever---numerical errors will eventually cause divergence---and therefore $\tau_\lambda$ gives us a metric of a reasonable timescale for prediction.

\subsection{Generative Models}
The dynamical models used for data generation have been chosen for their applicability to certain use cases as well as their ability to test particular abilities of reservoir computing. The Rossler system and Colpitt's oscillator are both 3 dimensional systems with long time scale dynamics---\ie, $\tau_\lambda$ large, slow error growth rates. The Colpitt's oscillator is particularly interesting because its dynamical equations of motion include an exponential term, making it the most nonlinear of the models studied here. The degree of nonlinearity of the dynamics can create greater difficulty for prediction with a linear readout operator. The Lorenz attractor (L63) is another 3 dimensional system but with shorter time scale dynamics than Rossler or Collpitts. The Lorenz 1996 system can be defined with arbitrary size so we can use it to look at how our RC scales with system dimension. Finally the Climate Lorenz Model (CL63) is composed of three L63 systems coupled together with multi-timescale dynamics.  RCs tend to focus on a particular time scale \cite{luk09} so it is important to study multi-timescale dynamics for certain applications such as in a weather prediction context \cite{kalnay2002, pena04}. See the appendix for equations and parameters.

\begin{table}[!htpb]
\centering
\begin{tabular}{||c c c||} 
 \hline
 Dataset & Largest LE $\lambda_1$ & Time Scale $\tau_\lambda = 1/\lambda_1$\\ [0.5ex] 
 \hline\hline
 Rossler & 0.065 & 15.4 \\ 
 \hline
 Colpitts & 0.07 & 14.3 \\
 \hline
 L63 & 0.9 & 1.1 \\
 \hline
 L96-5D & 0.4 & 2.5 \\
 \hline
 L96-10D & 1.1 & 0.9 \\ 
 \hline
 CL63 & 0.9 & 1.1\\[1ex] 
 \hline
\end{tabular} \caption{Dataset timescales. All dynamical systems we consider are made dimensionless, so the ``time'' here is number of dimensionless steps into the future.}
\end{table}

\subsection{Forecast Testing Metric}
A common ML benchmark for time series prediction is the one-step prediction mean square error Eq.\eqref{eq: loss_micro}. Indeed this is exactly how we train the micro-scale parameters in $\Wout$ to obtain the linear map from $\u(t) \rightarrow \r(t) \rightarrow \u(t+1)$. As a method for testing, however, one-step prediction is not an appropriate metric; it overemphasizes the high frequency modes in the data at the expense of the long-term stability of the RC forecast.

As shown by Platt et.al.,\cite{platt21}, the metric that best represents the ideal behavior of the RC---good short-term predictions coupled with ``physical'' (the forecast RC stays on the synchronization manifold $\M$) long-term forecasts---is best represented by the reproduction of the input system's LEs. In practice the LEs can be difficult to calculate from high-dimensional experimental data and computationally expensive for the RC. Therefore, a proxy for the reproduction of the LEs has in practice been long-term forecasts \cite{platt21}.

Our standard metric for forecast time is the valid prediction time (VPT) \cite{Vlachas20}. The VPT is the time $t$ when the accuracy of the forecast exceeds a given threshold. For example,
\begin{equation}
\rm RMSE(t) = \sqrt{\frac{1}{D} \sum_{i=1}^D \Big [\frac{u_i^f(t) - u_i(t)}{\sigma_i} \Big]^2} > \rm \epsilon, \label{eq: vpt}
\end{equation}
where $D$ is the system dimension, $\sigma$ is the long term standard deviation of the time series, $\epsilon$ is an arbitrary threshold, and $u^f$ is the RC forecast. For the results shown in subsequent section a total of 200 test initial conditions are used to report a distribution of VPTs with $\epsilon$ set arbitrarily to $0.3$.  \cite{Vlachas20} use $\epsilon \sim 0.5$ in their reported results.

\section{RC Training} \label{sec: training}
Training the RC model can be separated into training the macro-scale parameters---\eg, spectral radius, leak rate---and the micro-scale parameters \ie, the elements of $\Wout$. Much of the early guidance on RC training \cite{Lukoševičius12} included advice on the intuition behind hand-tuning the macro parameters for a particular problem as hyperparameters. For instance, a widely cited guideline is that the spectral radius must be below 1---see section \ref{sec: stability}---but that is based on many assumptions that do not necessarily hold for a particular RC application. Modern optimization algorithms and increased computational power have decreased the need to hand-tune all of these macro parameters as hyperparameters. Instead, they can be incorporated directly into the objective function for training via optimization. For intuition on the effect of these parameters on the RC we refer the reader to Lukoševičius \cite{Lukoševičius12}.

We employ a two step optimization approach, described by \cite{Griffith19, penny2021} where one first fixes the macro-scale parameters, trains the micro-scale parameters through linear regression Eq.\eqref{eq: sol_linear_reg} and then tests the completed RC on a series of long forecasts. For the macro loss function we have had success using the scaled error
\begin{equation*}
    \mathscr{L}_{\rm macro} = \sum_{i=1}^M \sum_{t=t_i}^{t_f} \norm{u^f(t) - u(t)}^2 \exp{-\frac{t - t_i}{t_f - t_i}}, \label{eq: global_loss}
\end{equation*}
with M being the number of forecasts used for comparison. We reiterate that the training for the macro-scale parameters does not use the one-step-ahead prediction error because that is not a good proxy for the reproduction of the LE spectrum. Penny et.al., \cite{penny2021} (Fig.12) examined the effect of increasing numbers of $M$ on the cost function landscape and showed that increasing $M$ smooths out the loss function landscape and enables better convergence to the global minimum. However, a compromise is necessary as increasing $M$ will substantially increase computational complexity. We have empirically found that using 15-20 or so forecasts is generally enough for these simple models, provided that those forecasts are made at statistically independent points. Ideally the data would well sample the input system attractor, but in practice that cannot always be the case.

\subsection{Bayesian Optimization}
Rather than hand-tune each macro-scale parameter as a hyperparameter, we suggest instead using an optimization procedure such as the two step optimization routine described above. While any global search algorithm can be used \eg, differential evolution, we use a Bayesian optimization technique using surrogate modeling. Bayesian optimization is a protocol for optimizing expensive nonlinear functions that works by “minimiz[ing] the expected deviation from the extremum” of a target loss function \cite{Mockus75}. We have found success with the efficient global optimization algorithm of Jones et.al., \cite{Jones98} as implemented by \cite{Bouhlel19}. This algorithm evaluates the loss function over a number of sampled points and then a Gaussian process regression is fit to the surface produced. One can then optimize over this interpolated ``surrogate'' model to identify promising regions of parameter space and then iterate by reevaluating the loss function in these regions. Given that we need only train a few macro-scale parameters, this surrogate optimization technique renders the optimization problem computationally tractable.

\begin{figure}[!htpb]
    \centering
    \includegraphics[width=0.45\textwidth]{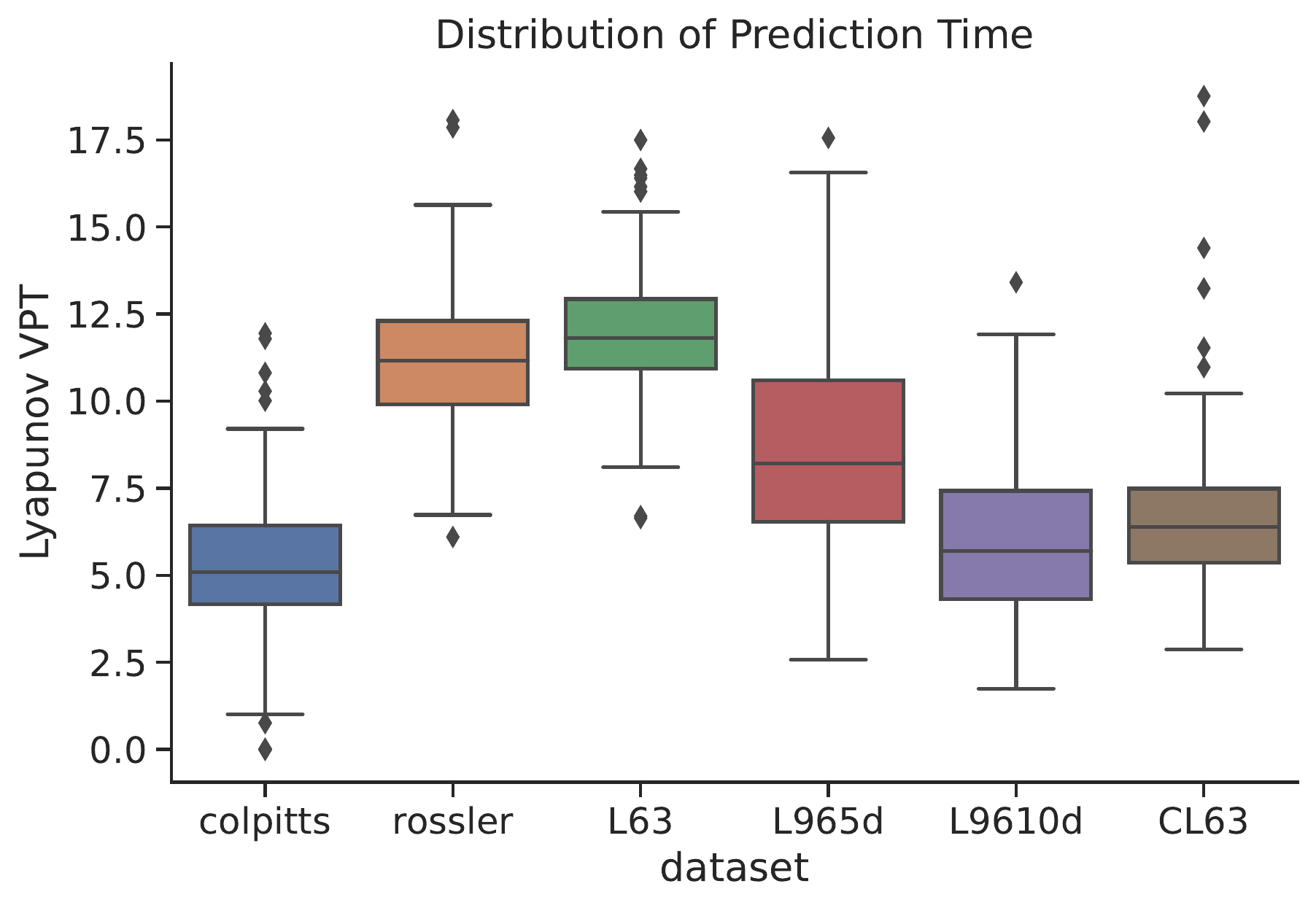}
    \includegraphics[width = 0.45\textwidth]{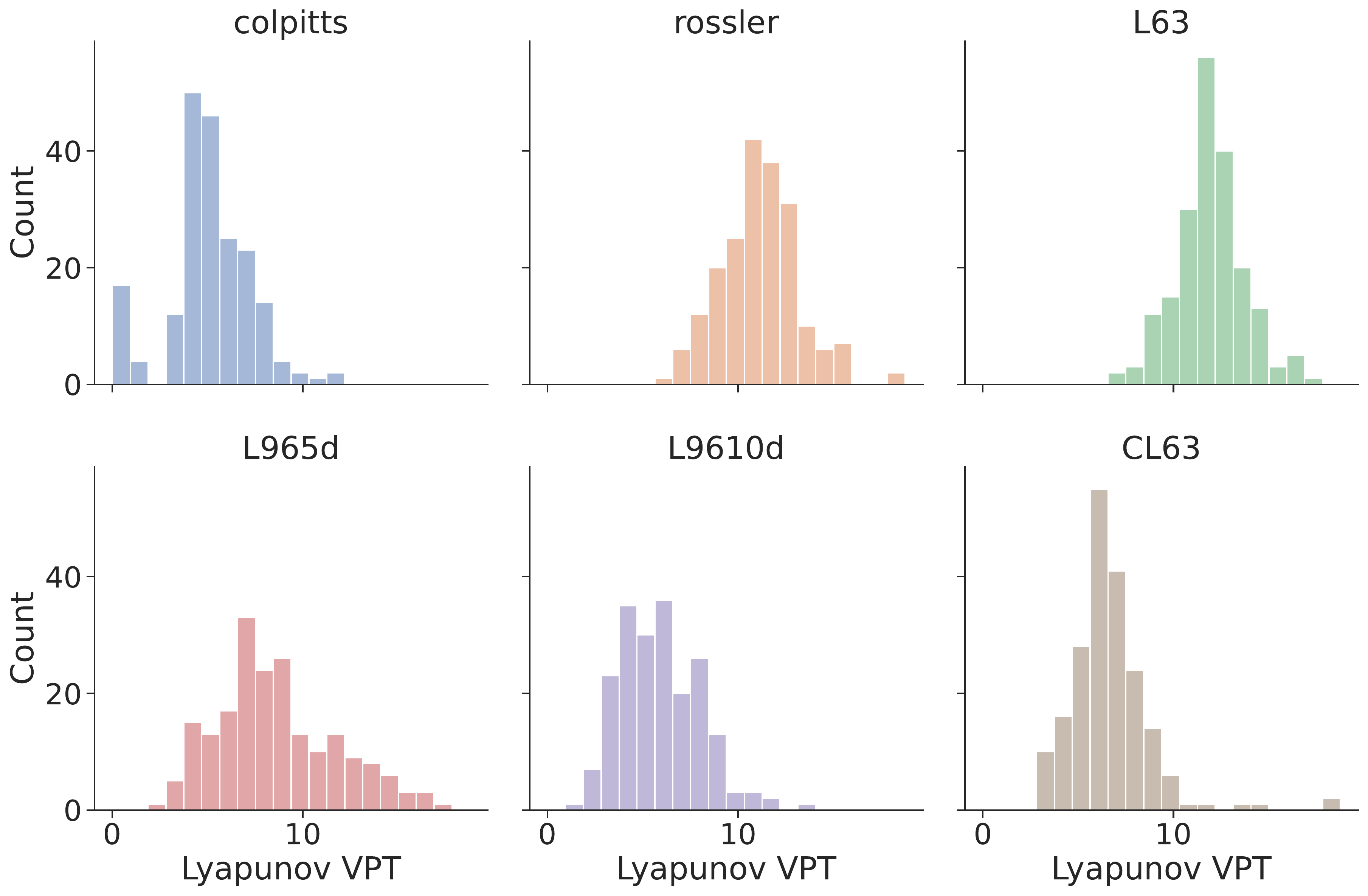}
    \caption{\textbf{(left)} Valid prediction time for the best fit parameters of the RC over different datasets. Note the distribution of forecast times. The RC predicts well on all the models given to it. Results are shown for $N=2000$, results for $N=250$ are shown in Fig.(\ref{fig: res_dim}).\\  \textbf{(right)}  Histogram view of the distribution of VPT for the different models. The distributions are close to Gaussian but with heavier tails. The outliers correlate with the FTLEs of the input system. When the FTLEs are low negative values, meaning the input data is very stable, then prediction times can be extremely long and vice versa.  While some of the variability in prediction time is caused by randomness and training in the RC, other variability is intrinsic to the dynamics we are attempting to predict.}
    \label{fig: best}
\end{figure}

With these algorithms, one can jettison the hand-tuning of hyperparameters and simply optimize all the macro-scale RC parameters for a given problem. This procedure produces good performance for a wide range of simple dynamical models Fig.(\ref{fig: best}) in a reasonable amount of time.

We will now go into greater detail on the choices that must be made outside of the optimization routine (by definition hyperparameters) and how they can affect the solution to the RC problem. The experiments are conducted by fixing/varying a particular hyperparameter or macro-scale parameter from table 1 and then optimizing over the rest of the macro-scale parameters. It will be stated if we are not optimizing over the un-varied parameters.

\subsection{Reservoir Dimension}
\label{ssec:reservoir_dimension}

\begin{figure}[!htpb]
    \centering
    \includegraphics[width = 0.45\textwidth]{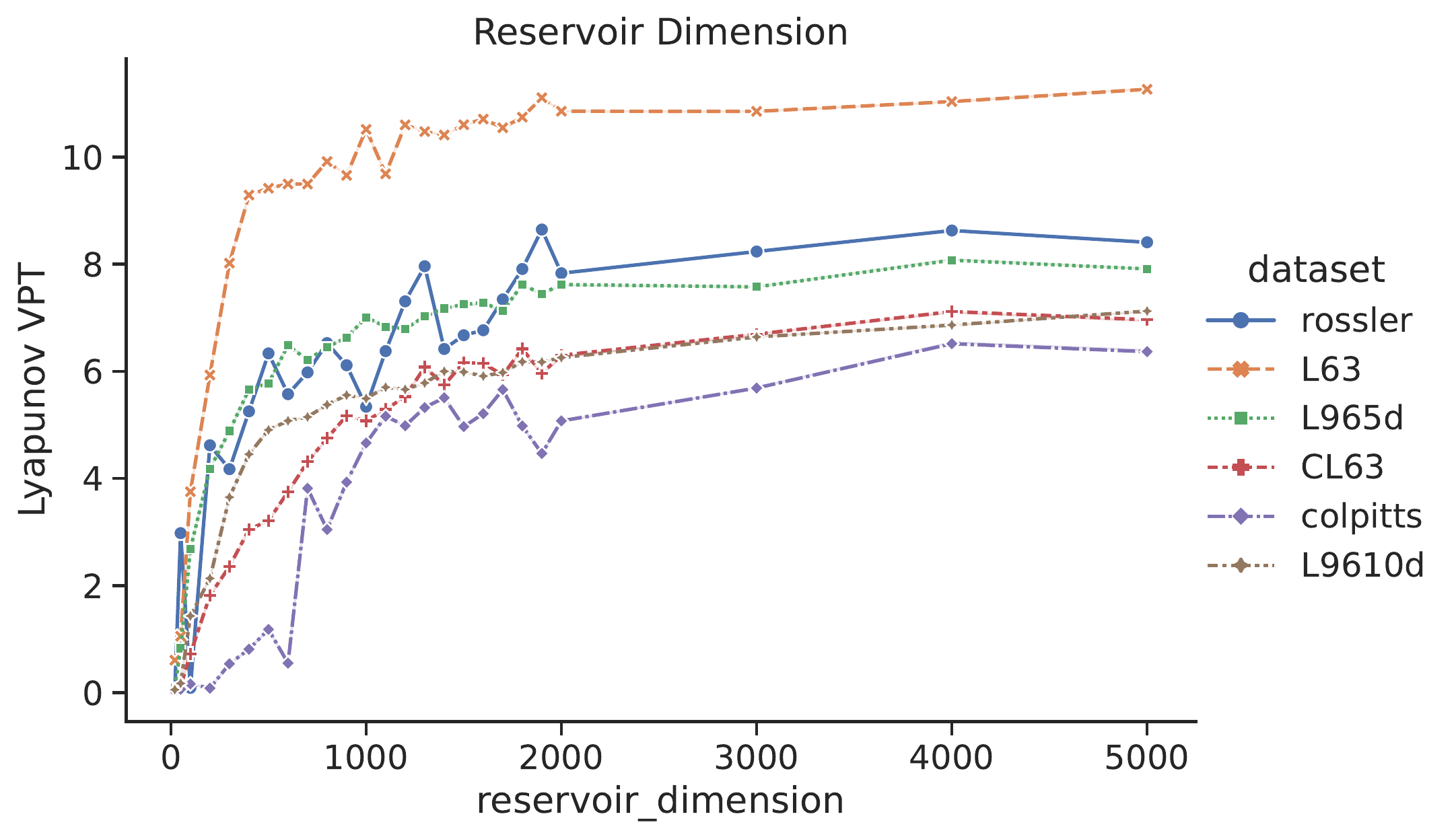}
    \includegraphics[width = 0.5\textwidth]{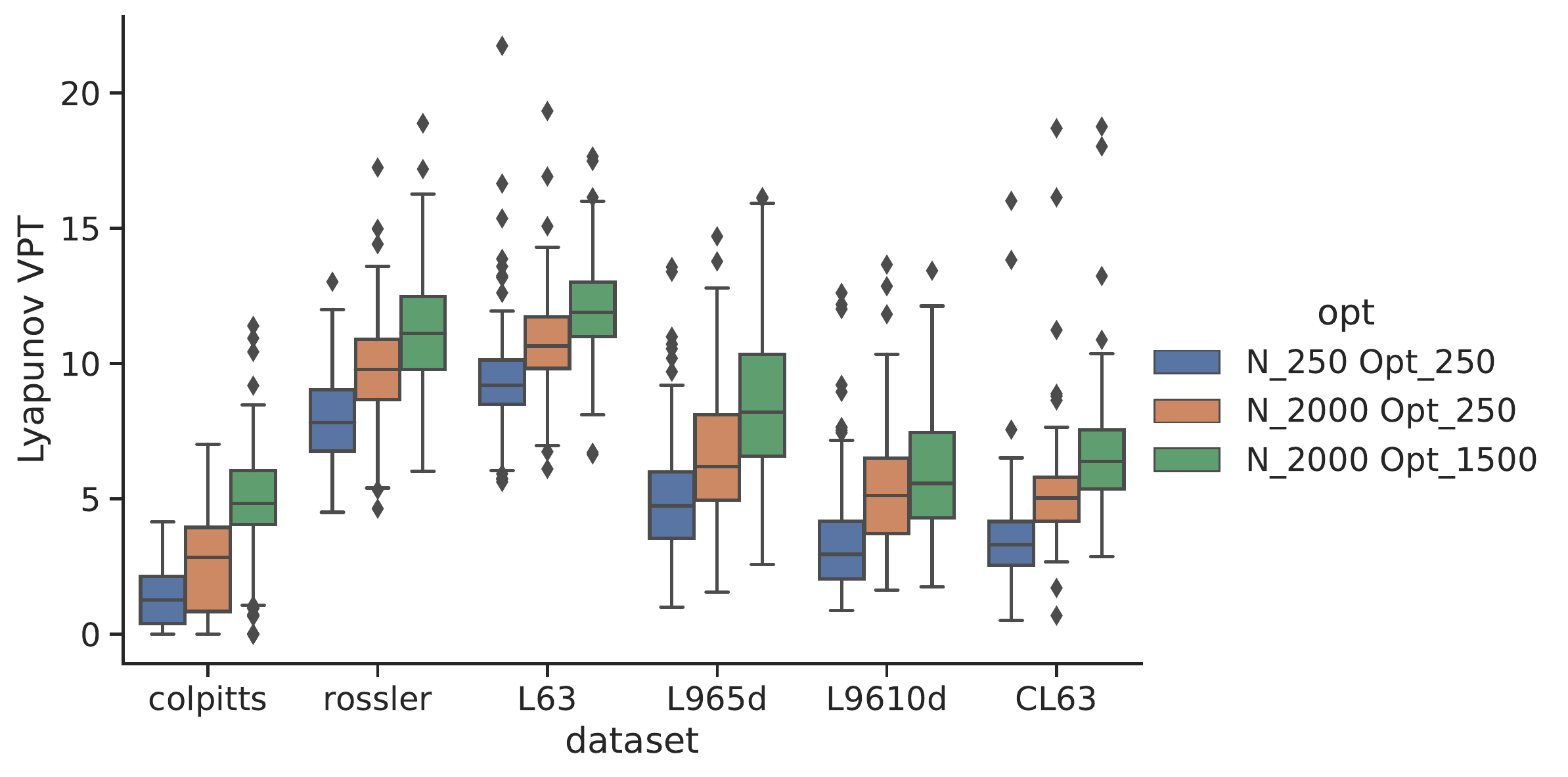}
    \caption{\textbf{(left)} The effect of increasing reservoir dimension without reoptimization of the global parameters. RCs optimized for $N$ = 2000.\\  \textbf{(right)} VPT for a 3 different RCs 1) small $N = 250$ RC optimized at $N = 250$ 2) an RC with identical global parameters to 1, but with a dimension scaled up to $N = 2000$ 3) an RC with $N = 2000$ optimized for $N = 1500$.}
    \label{fig: res_dim}
\end{figure}

The reservoir dimension $N$ of the RC is related to the memory capacity of the network \cite{Jaeger02, Lukoševičius12}.  Larger reservoirs do tend to produce more accurate forecasts with longer VPTs than smaller networks. However, there are diminishing returns where ever larger reservoirs are needed for ever smaller improvements in predictability. For example, consider the L63 system in Fig.(\ref{fig: res_dim}). Using a reservoir size of just $250$ nodes, we see a VPT only a couple Lyapunov time scales less than when using a reservoir of size $2000$. Therefore if our application only requires a few Lyapunov time scales of prediction skill, then it is possible to use a much smaller and more computationally efficient RC.

Because optimization routines are far more computationally intensive than forecasting with the RC, it is practical to ask whether one can optimize/train using a smaller reservoir and then scale up to larger reservoir to generate a forecast. The answer is that one can indeed train a small reservoir and then use those same macro-scale parameters with a larger reservoir to see substantial improvement in prediction skill. This strategy was used by \cite{penny2021}. However, an RC optimized at the larger size will tend to produce more accurate forecasts. If computational resources allow, it is appropriate to perform a final optimization using the reservoir size that will be used for the production RC. The left side of Fig.(\ref{fig: res_dim}) shows the effect of increasing the reservoir dimension without re-optimizing parameters. The RC performance tends to saturate as the reservoir dimension increases.

\subsection{Spinup Time}
The spinup time is the time it takes for a trained RC to converge from its initial condition (usually either set to zero or randomly chosen) onto the synchronization manifold to which it is driven by the input data. The number of steps needed for spinup is related to:
\begin{enumerate}
\item How far the initial condition of the RC is from the synchronization manifold.
\item The CLEs of the generalized synchronization manifold; these are related to the ``echo state property'' and the spectral radius of the adjacency matrix $\A$.
\item The RC approaches the synchronization manifold asymptotically so while the error tolerance is also important, it probably does not need to converge to within $10^{-12}$. Convergence during spinup and error growth during prediction are both governed by $\delta x(t) \approx \delta x_0 e^{\lambda t}$ with $\lambda$ giving the negative CLEs for spinup or the positive LEs for forecasting.
\end{enumerate}

\begin{figure}[!htpb]
    \centering
    \includegraphics[width=0.5\textwidth]{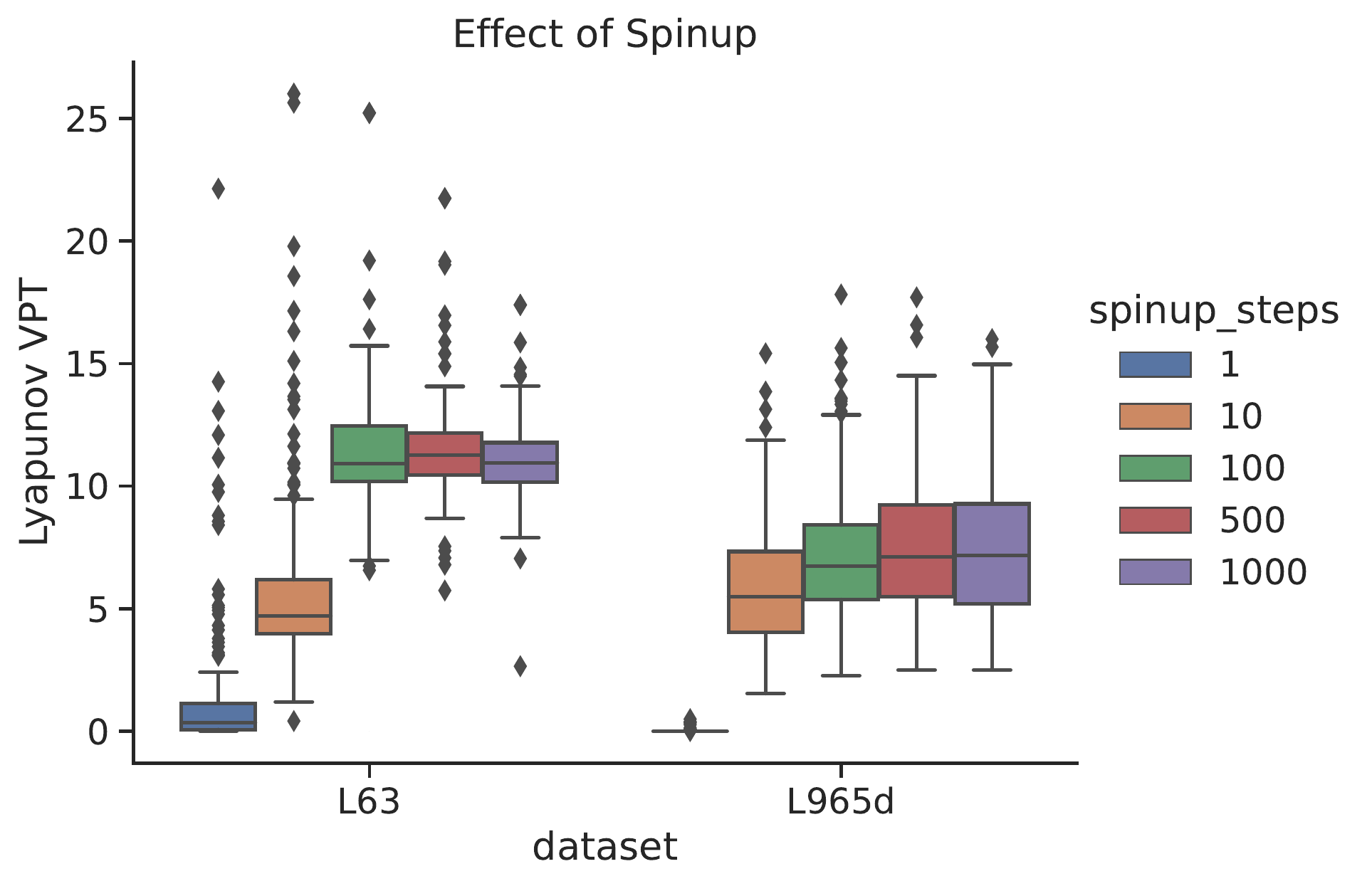}
    \caption{Effect of setting the number of spinup steps on the VPT of the RC for L63 and L96-5D.  Longer spinup times reduce the initial error of the forecast as the RC and input data synchronize together.}
    \label{fig: spinup}
\end{figure}

In practice it is easy to tell empirically how much time is needed---see Fig.(\ref{fig: spinup})---usually for these small models only $\mathcal{O}(10)$ time steps are necessary. For larger systems the spinup time can be significantly higher.

\subsection{Input Bias}
\begin{figure}[!htpb]
    \centering
    \includegraphics[width = 0.75\textwidth]{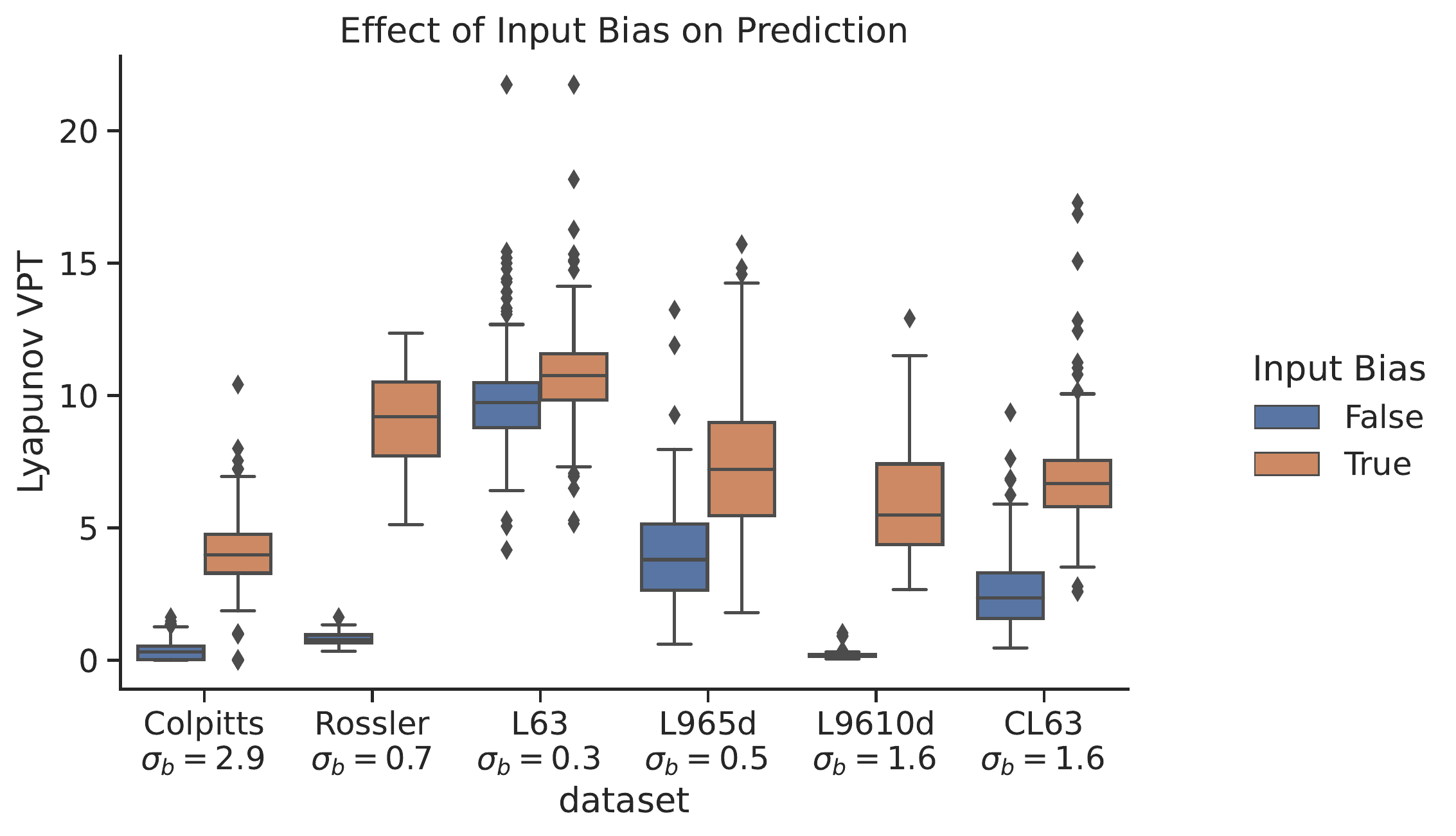}
    \caption{There are two sets of parameters: 1) no input bias 2) input bias. Both sets of parameters are optimized separately but with the same reservoir size. The readout is linear for both.  The input bias is a constant $\sigma_b$, which is optimized Eq.\eqref{eq: mapping_form}. $\sigma_b$ sets the nonlinear regime of the $\tanh$. When $\sigma_b = 0$ the optimization is not able to find any parameters that work for many of the models. However, when $\sigma_b$ is optimized all models produce accurate and reliable forecasts.}
    \label{fig: input_bias}
\end{figure}
\begin{figure}[!htpb]
    \centering
    \includegraphics[width = 0.6\textwidth]{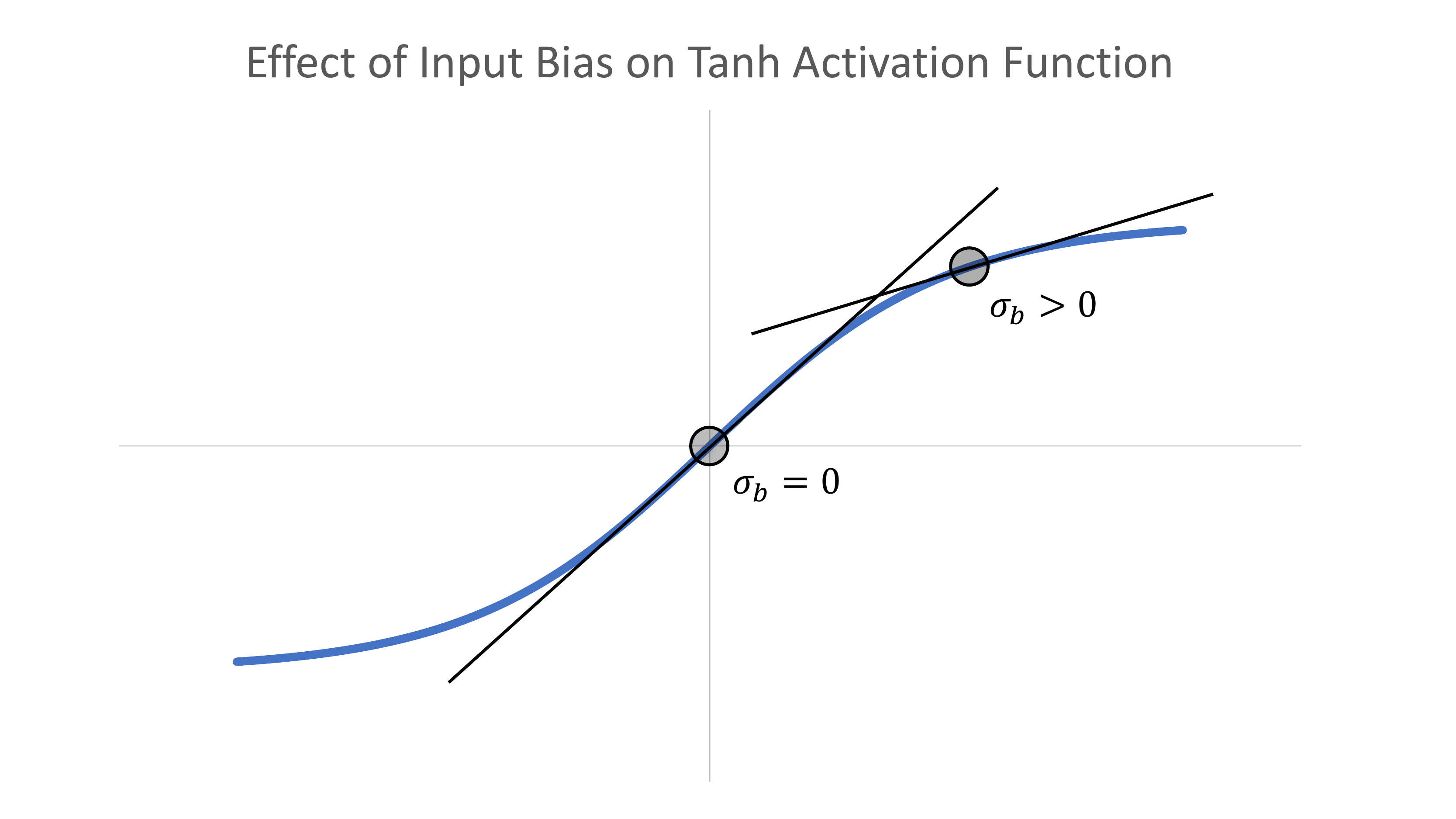}
    \caption{Effect of input bias $\sigma_b$ on tanh activation function. When $\sigma_b \approx 0$, the $\tanh(\cdot)$ activation function operates mostly around $0$ in the ``linear'' regime of the activation function. A nonzero $\sigma_b$ will push the operating point to the nonlinear regime of the $\tanh(\cdot)$ activation function, possibly giving more expressive results. The bias also affects the sensitivity of the RC to input. Around the origin the RC has maximal gain, while a high bias term will reduce the gain shown in the figure by the slope of the lines.}
    \label{fig: tanh}
\end{figure}

The input bias $\sigma_b$ can have a significant impact on RC prediction skill, but it has consistently been left out of most published RC studies focused on forecasting chaotic dynamics. We see in Fig.(\ref{fig: input_bias}) that if we do not include this term in the RC when optimizing macro-scale parameters then for almost all models the forecast skill is negligible. The input bias sets the extent of the `nonlinearity' in the reservoir Fig.(\ref{fig: tanh}); $\sigma_b=0$ corresponds to the reservoir operating mostly in the linear part of the tanh function around 0, while a high bias term corresponds to the RC operation being in a more nonlinear part of the regime.

\subsection{Readout Functions: Linear, Biased, Quadratic} \label{sec: readout}
\begin{figure}[!htpb]
    \centering
    \includegraphics[width=0.6\textwidth]{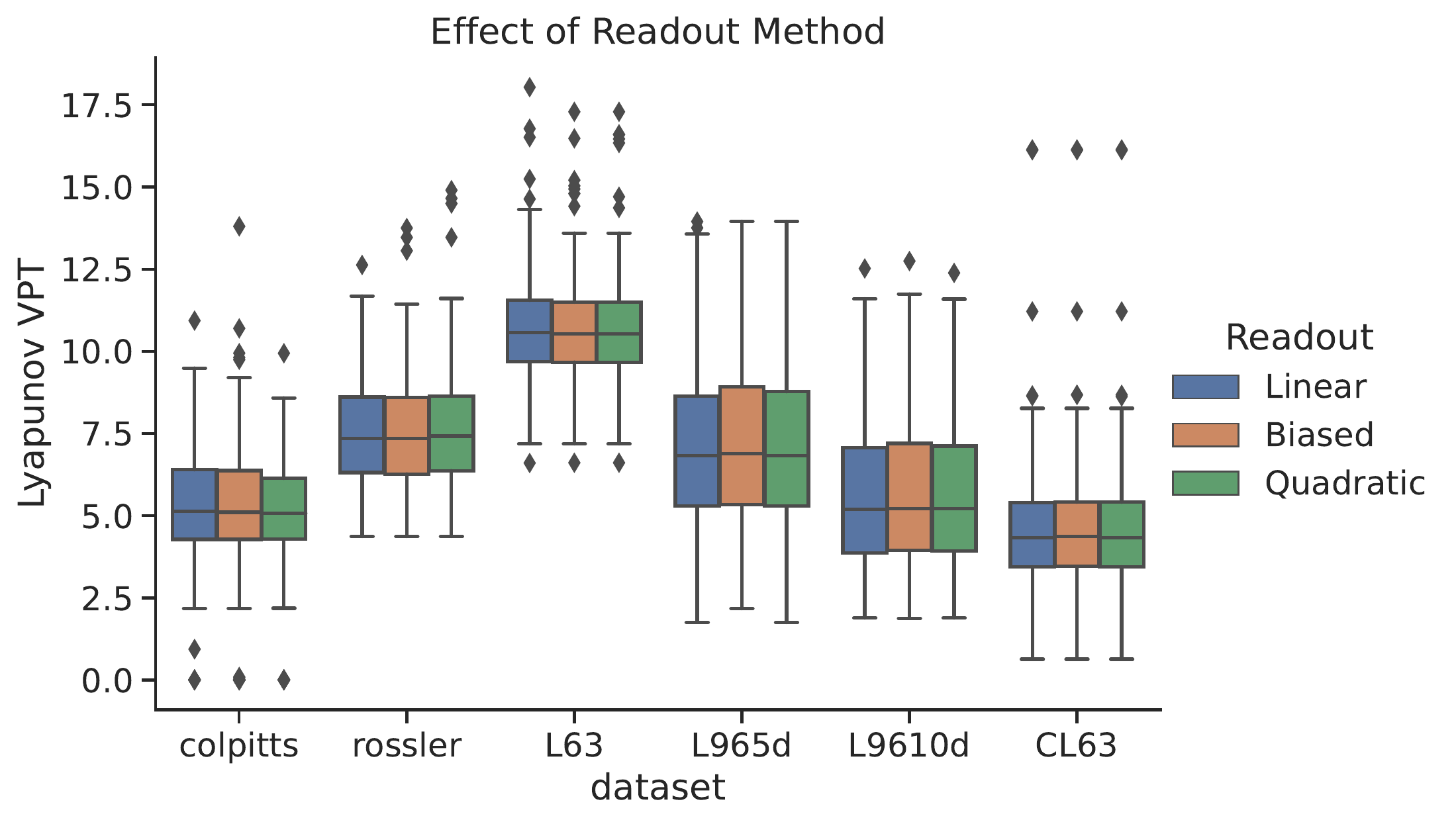}
    \caption{Predictions for 3 different kinds of readout methods.  Linear: $\hat \u(t) = \Wout \r(t)$, Biased: $\hat \u(t) = \Wout [\r(t), \u(t-1)]$, Quadratic: $\hat \u(t) = \Wout [\r(t), \r^2(t)]$.  }
    \label{fig: readout}
\end{figure}

We generalize the readout $\Wout \r$ to $\Wout Q(\r)$ for some function $Q$. An example in the literature is by Lu et.al., \cite{Lu18} who use a quadratic readout function $Q(\r) = \mqty[\r(t), \r^2(t)]$. As another example, \cite{Lukoševičius12} recommended a biased output where $Q(\r, \u) = \mqty[\r(t), \u(t-1)]$ so the previous input is fed directly into $\Wout$. We have tested a number of options and found that the readout does not have much impact on the quality of the forecasts made by the RC Fig.(\ref{fig: readout}) as long as the parameters are well optimized and the input bias $\sigma_b$ is not set to 0. While these results seemingly contradict those reported and used in the previously cited studies, we note that when the bias term is not included---see Fig.(\ref{fig: input_bias})---then changing the readout layer does indeed improve the predictions. We speculate that the readout in this case is injecting a certain amount of nonlinearity, essential for good predictions, into a system that is operating around the linear regime of the activation function.

\subsection{Amount of Training Data}

\begin{figure}[!htpb]
    \centering
    \includegraphics[width=0.45\textwidth]{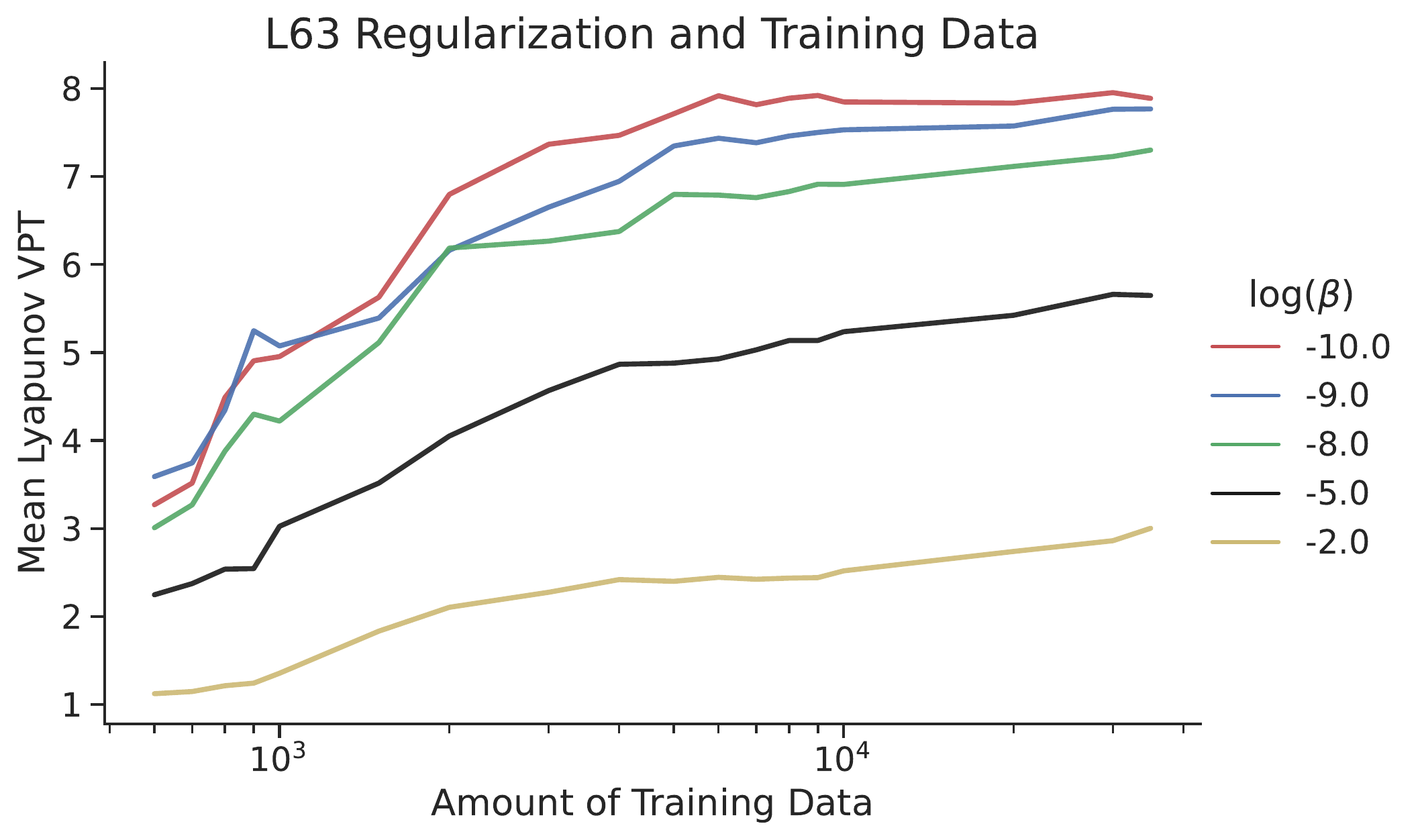}
    \includegraphics[width=0.45\textwidth]{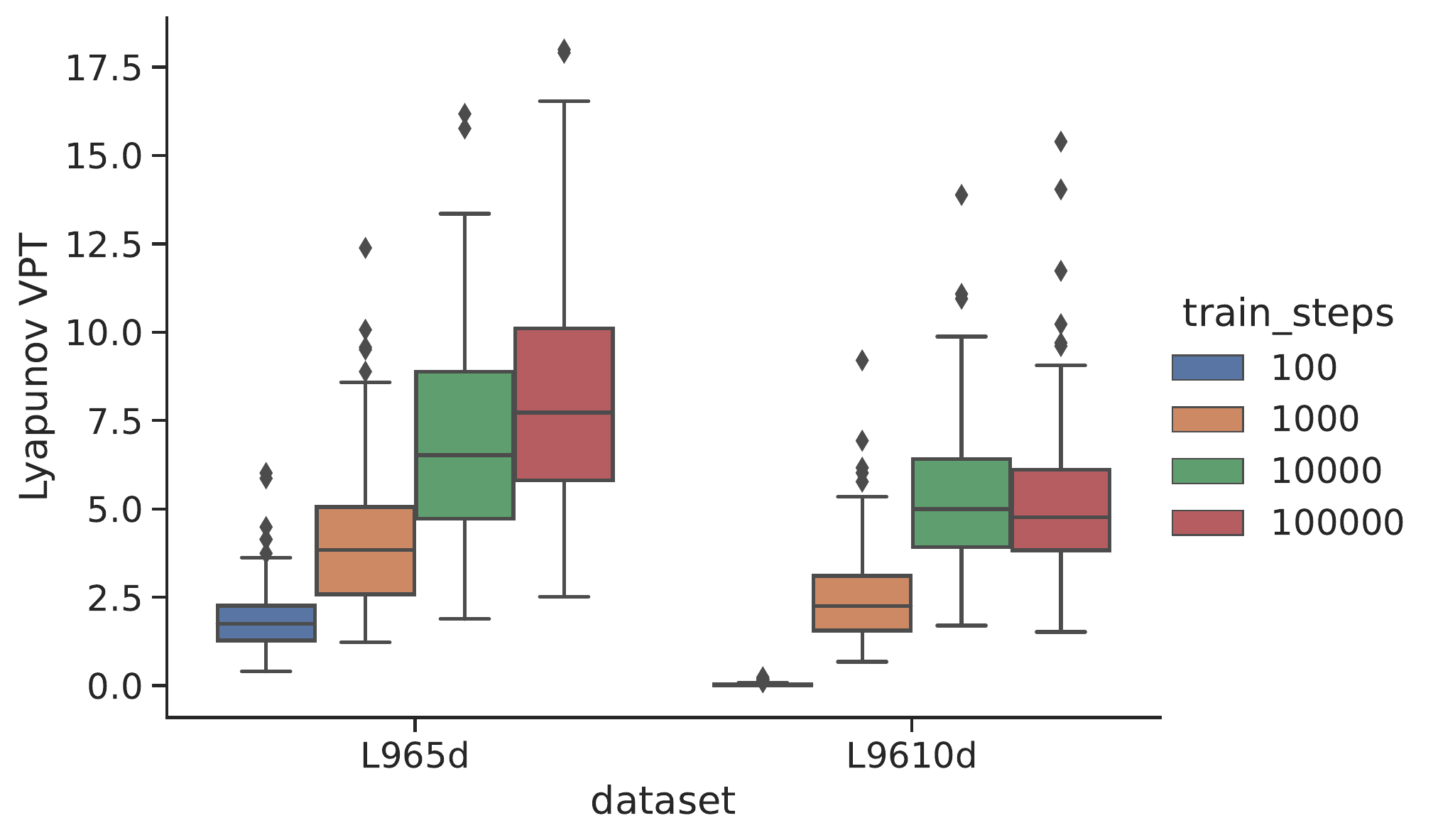}
    \caption{(left) L63, RC N=250 for different values of regularization without reoptimizing all the parameters. Diminishing returns with increasing amounts of data. (right) Here we re-optimize the RC for each amount of training data. We still see a clear reduction in the rate of increasing skill as the amount of training data increases.}
    \label{fig: reg_training}
\end{figure}

The data are used differently in the two steps of the training routine---see Eq.\eqref{eq: global_loss}. The data should be separated into the $M$ long forecasts that are used to compute the loss $\mathscr{L}_{\rm macro}$ and the data used to train $\Wout$ in Eq.\eqref{eq: sol_linear_reg}. More and longer test forecasts for the global loss will help guarantee 1) the generalizability of the RC to segments of the data the RC might not have seen, 2) stability of the trained RC, and 3) the optimality of the global parameters found during training. The amount of data that is used for the micro-scale loss function governs the accuracy of the linear map $\Wout$. Experiments indicate diminishing returns, where increasing the size of the training data leads to diminishing improvements in forecast skill Fig.(\ref{fig: reg_training}).

\subsection{Normalization}
\begin{figure}[!htpb]
    \centering
    \includegraphics[width = 0.7\textwidth]{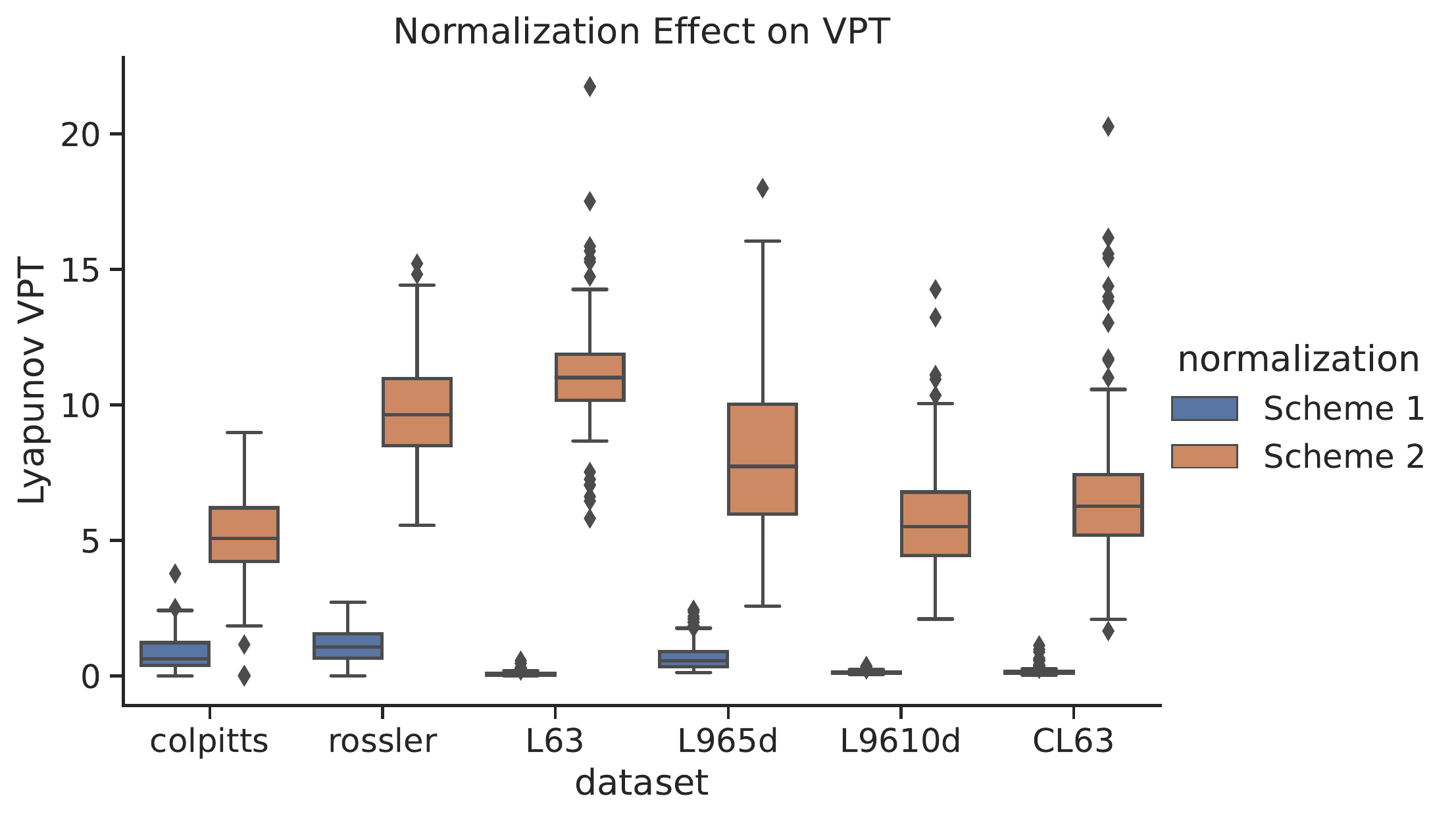}
    \caption{Results from normalizing by the two different schemes. The results from scheme 1 can actually be improved slightly by adding a small $\sim 1-2\%$ amount noise to the data---a phenomenon described as well in \cite{Vlachas20}---but the improvement is marginal.  While we are not advocating for a particular normalization scheme, particularly for already nondimensionalized data, it is important to keep in mind that there is essential information in the relationships between the time series that can be destroyed by introducing normalization.}
    \label{fig: norm_results}
\end{figure}

Normalization is a standard practice in ML. Applying normalization is usually among the first suggestions to anyone analyzing a new dataset \cite{Goodfellow16}. We must emphasize, however, that when forecasting dynamical systems the standard methods for normalizing datasets can have deleterious effects. To illustrate, we introduce two normalization schemes,

\begin{enumerate}
    \item Normalize each variable separately by subtracting the mean and dividing by the standard deviation
    \begin{equation*}
        u^{\rm norm}_i = \frac{u_i - \rm{mean}(u_i)}{\rm{std}(u_i)};\ i \in \mqty[1, 2, \ldots, D].
    \end{equation*}
    \item Normalize the variables by the joint mean and max/min of the variables
    \begin{equation*}
        u^{\rm norm}_i = \frac{u_i - \rm{mean}(\u)}{\max{\u} - \min{\u}}\ i \in \mqty[1, 2, \ldots, D].
    \end{equation*}
\end{enumerate}
We assume that the variables in the dynamical system have been nondimensionalized. The statistical functions mean, max, and min are calculated from all the training data and thus represent ``climatological'' statistics. The VPTs resulting from both normalization schemes are shown in Fig.(\ref{fig: norm_results}).

As shown in Figure \ref{fig: norm_results}, normalization scheme 2 vastly outperforms scheme 1. Our hypothesis is that normalizing each variable separately (particularly subtracting the mean) destroys the cross-variable information necessary to reconstruct the relationships in the dynamical system. There is a significant amount of information stored in the relationships between the variables as well as their magnitudes. The second scheme compensates for the normalization via $\sigma$ and $\sigma_b$ to preserve all the information in the data while the first scheme destroys that information.

It is possible that we could recover the prediction skill by a skillful selection of $\Win$. If we optimized the inputs ($\sigma$) for each column of $\Win$ separately as well as $\sigma_b$, that would help the RC to compensate for the normalization. However, this introduces additional global parameters that must be optimized.

Generally, we recommend not normalizing the variables when the data is nondimensionalized, as is the case for the dynamical models examined here. We note that \cite{penny2021} used no normalization and still produced successful predictions with the RC models. In a realistic setting, however, when the data is collected from sensors and there are many different unit scales, one has to do some kind of normalization in order for the methods to be numerically stable. We discuss this point further with a simple example in appendix \ref{app: malkus}. Because we have shown that normalization can be detrimental when using data to forecast dynamical systems, we emphasize that care must be taken when normalizing any data prior to training.

\subsection{Effect of Noise}

\begin{figure}[!htpb]
    \centering
    \includegraphics[width = 0.55\textwidth]{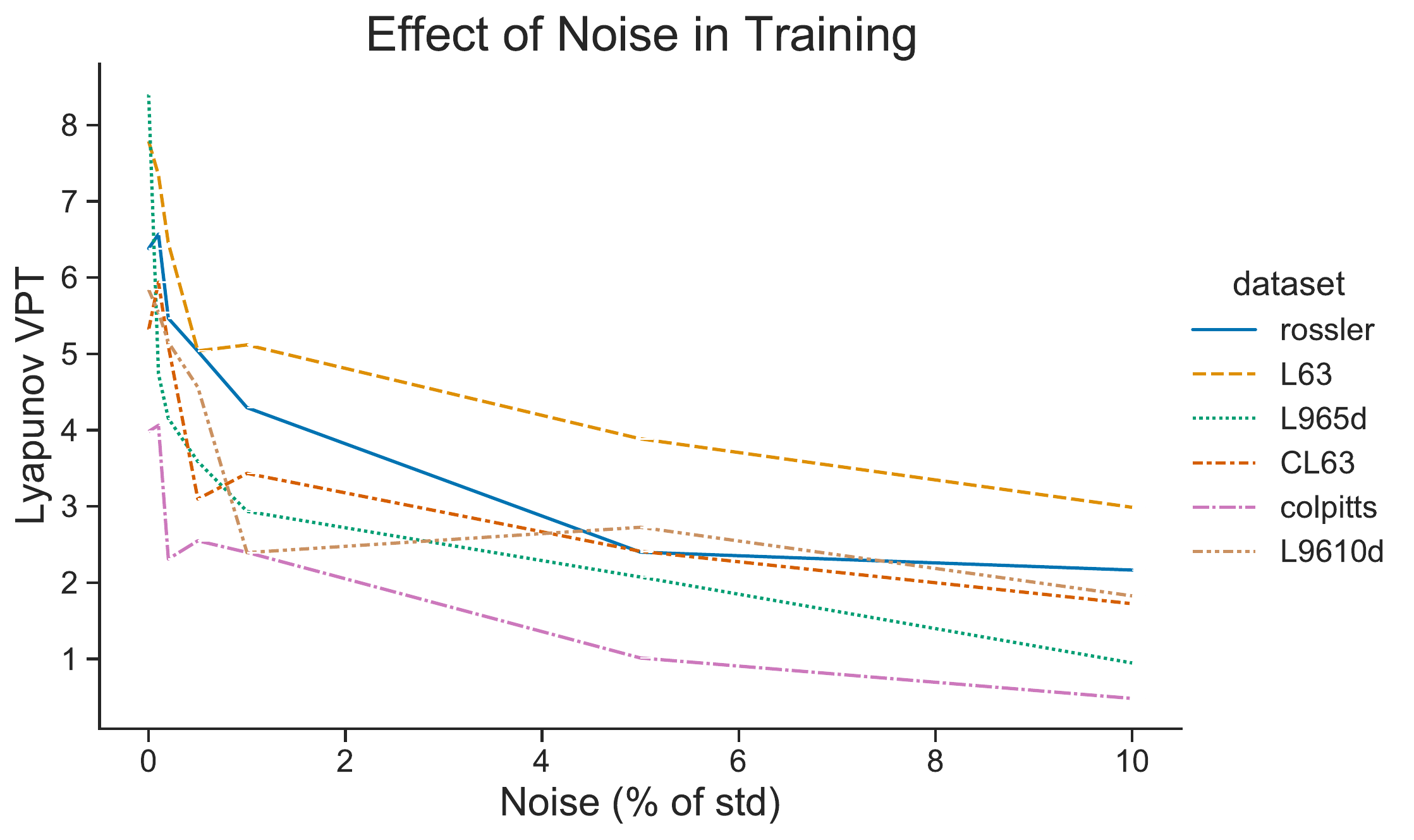}
    \caption{Mean VPT as a function of additive Gaussian noise expressed as a percentage of the long term standard deviation of each dynamical variable. The parameters were optimized for each noise level.}
    \label{fig: additive_noise}
\end{figure}

The RC is quite sensitive to additive noise Fig.(\ref{fig: additive_noise}). There is a sharp decrease in prediction quality with even a small amount of noise, meaning that any slight perturbation to the training of $\Wout$ causes the errors to multiply rapidly. This is consistent with our observation that even casting double precision floating point numbers in $\Wout$ to single precision can cause similar degradation.

We must address a phenomenon reported by \cite{Vlachas20} that sometimes a small amount of noise can help predictions. This observation held true when we normalized the data using scheme 1 rather than scheme 2---see the previous section. We note, however, that even with the small observed increase in predictive skill, the VPT was still far below that reported with the scheme 2. While \cite{Vlachas20} speculated that the noise could regularize the data, we find this is not true in general.

\subsubsection{Sparse in Time}
\begin{figure}[!htpb]
    \centering
    \includegraphics[width = 0.7\textwidth]{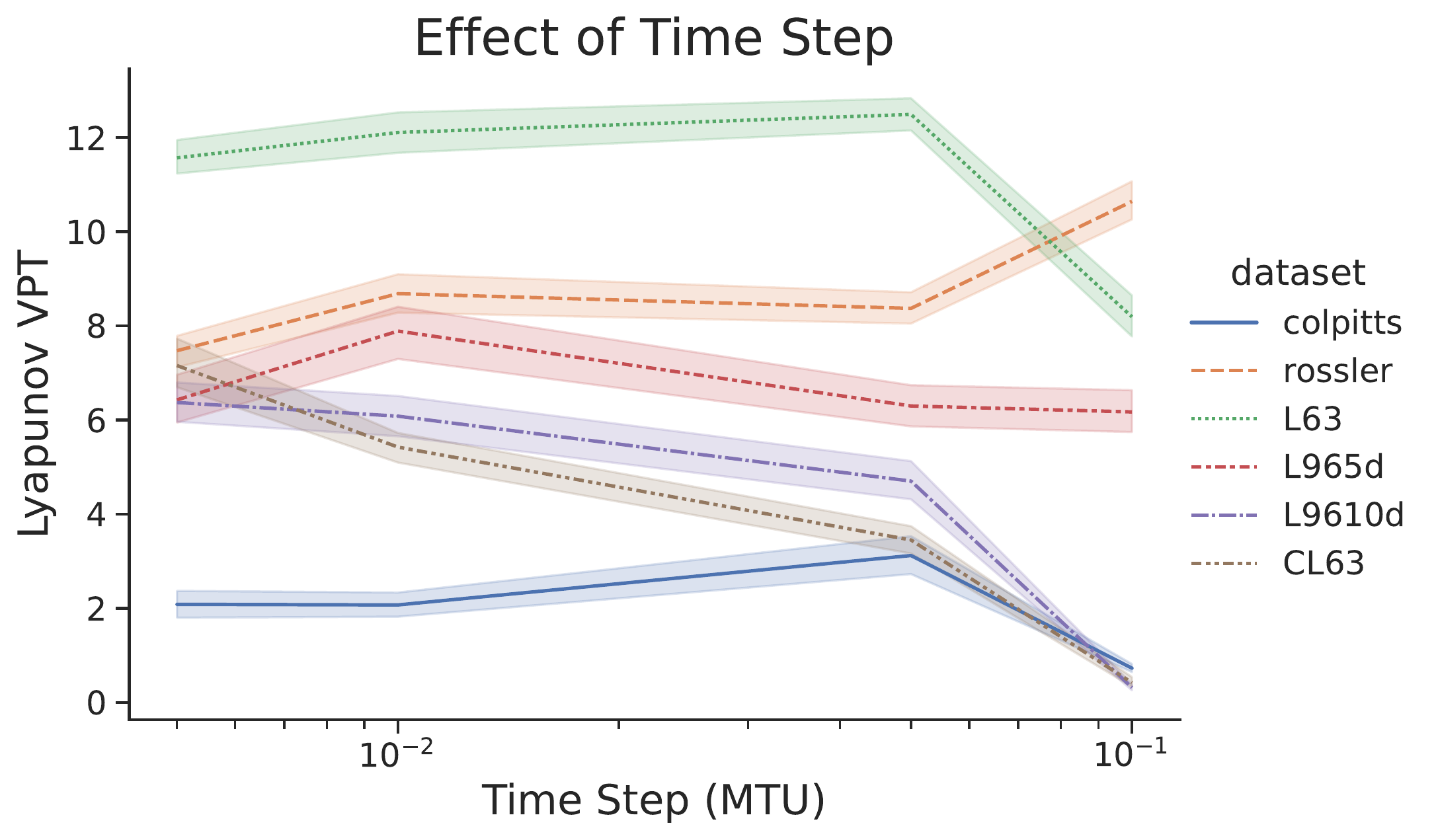}
    \caption{VPT as a function of the time step of the RC. The shading denotes the 95\% confidence interval for the mean of the distribution. Note the decrease in VPT as the time step increases for the L9610d and CL63 systems and increase for the Rossler system.}
    \label{fig: time step}
\end{figure}

RC can be quite sensitive to the time step of the input data \ie, how sparse the measurements are in time. In Fig.(\ref{fig: time step}) we tested the RC for different time steps $\Delta t$, reoptimizing the RC each time. The total time of the training data $T = L \Delta t$ for $L$ number of steps, did not change. We call attention to the result that finer training data does not always increase the prediction time of the RC and that certain models had an optimal time step. In practice one may want to interpolate the input data to try to match the optimal time step.

\subsection{Sparsity of the Adjacency Matrix $\A$}
The sparsity of the adjacency matrix in general makes very little difference to the predictive capability of the RC---this is described by \cite{Lukoševičius12} and matches our experience. The benefit of using a very sparse matrix comes from taking advantage of sparse matrix representations in scientific computing software packages to vastly increase computational speed. We have found a set value of 98\% or 99\% sparsity, $\rho_A = 0.01$ to be sufficient for all applications, provided the resulting matrix is full rank.

Of course there is a limit to this guideline---if the matrix is too sparse (\ie, has no nonzero values) then the RC will not work. In addition, when using small reservoirs of say less than $N$=$100$ then the randomness of the selection of $\A$ can have a significant impact on the quality of predictions \cite{luk09}. Slightly larger reservoirs seem to be more robust to random variations in connectivity.


\section{Scaling to Higher Dimensional Systems} \label{sec: scaling up}

\begin{figure}
\centering
\includegraphics[width=.6\textwidth]{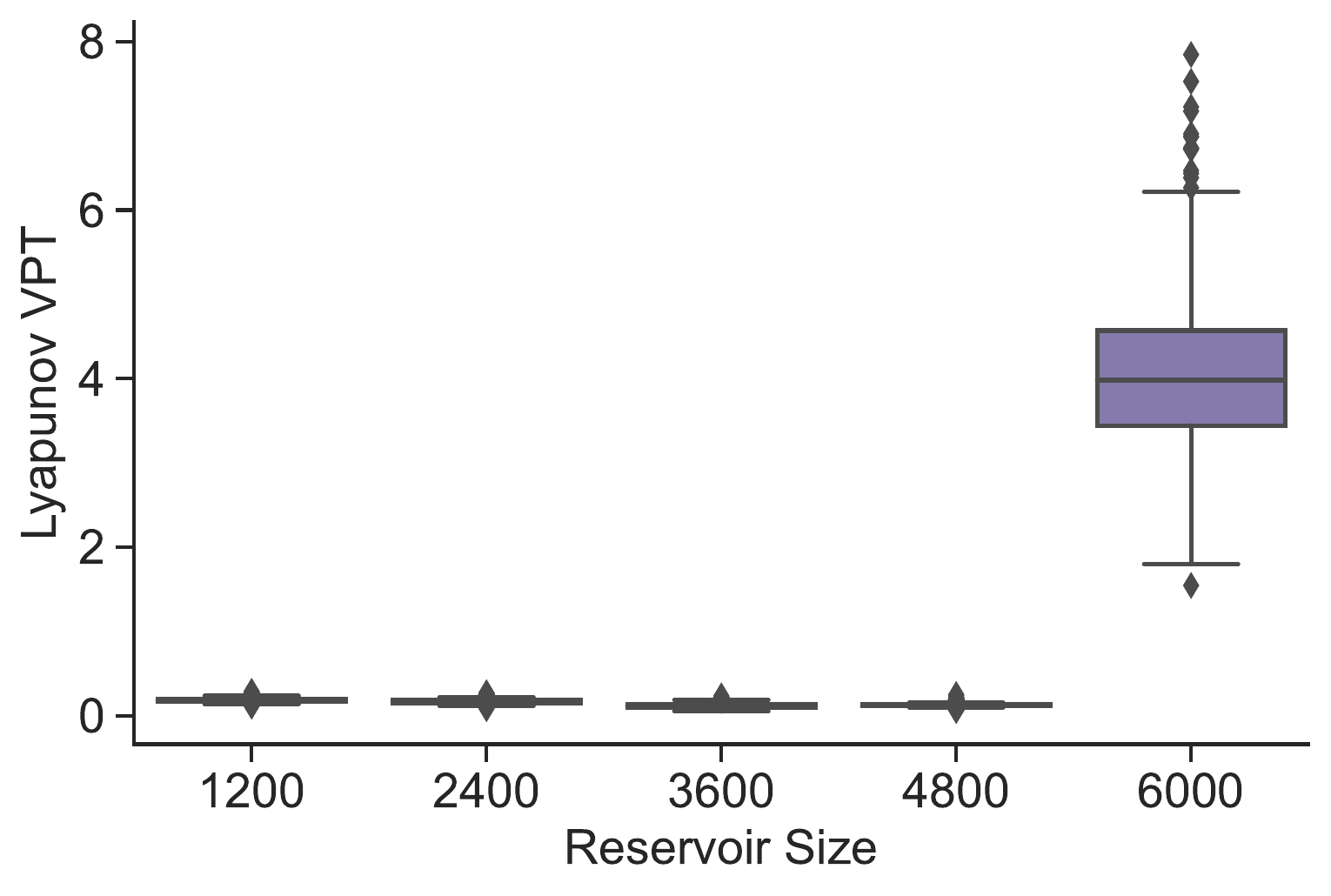}
\caption{VPT from predictions of the 40D Lorenz96 system with a single RC and no localization. Parameters are optimized for each reservoir size. Parameter values are given in Table \ref{tab:why_scale_up}.}
\label{fig:why_scale_up}
\end{figure}

Up to this point we have considered systems with a state dimension $D\leq 10$, and a single RC model has been sufficient to predict their time evolution.  However, systems with larger state spaces will require larger reservoirs to make adequate predictions. We highlight this situation in Figure \ref{fig:why_scale_up}, which shows the VPT of RC models with increasing reservoir dimension on the Lorenz96 system with 40 nodes. Note that we use a leading Lyapunov exponent of $\lambda_1\simeq 1.68$ to represent the timescale for this system following \cite{Vlachas20}. There is no prediction skill until a certain minimum reservoir size is attained,
somewhere between 4,800 and 6,000. For systems with even larger state spaces, the reservoir size will have to increase beyond the amount of Random Access Memory (RAM) available and parallelization schemes will have to be considered.

To address these issues, \cite{pathak18} introduced a parallelization strategy which has been used in numerous high dimensional applications, e.g.\ \cite{arcomano20,Vlachas20,penny2021}.
In this localization scheme, multiple reservoirs work semi-independently, each making predictions of a
subset of the system state. More precisely, the elements of the system state $\textbf{u}\in\mathbb{R}^{D}$ are split up into
$N_g$ groups, such that each group is assigned an individual RC model that predicts $N_\text{output}$ of the state vector nodes, with $D = N_g N_\text{output}$.
At each time step, each group receives $N_\text{halo}$ of the neighboring state vector values, such that a system with a single spatial dimension has an input dimension for each RC model of $N_\text{input} = 2 N_\text{halo} + N_\text{output}$.
With this architecture one must decide how to choose $N_\text{output}$ and $N_\text{halo}$. Our goal in this section is to explore these choices within the framework outlined in Sections \ref{sec: testing} \& \ref{sec: training}.

\subsection{Varying the halo size and output dimension}
\label{ssec:halo_and_output}

\begin{figure}
\centering
\includegraphics[width=\textwidth]{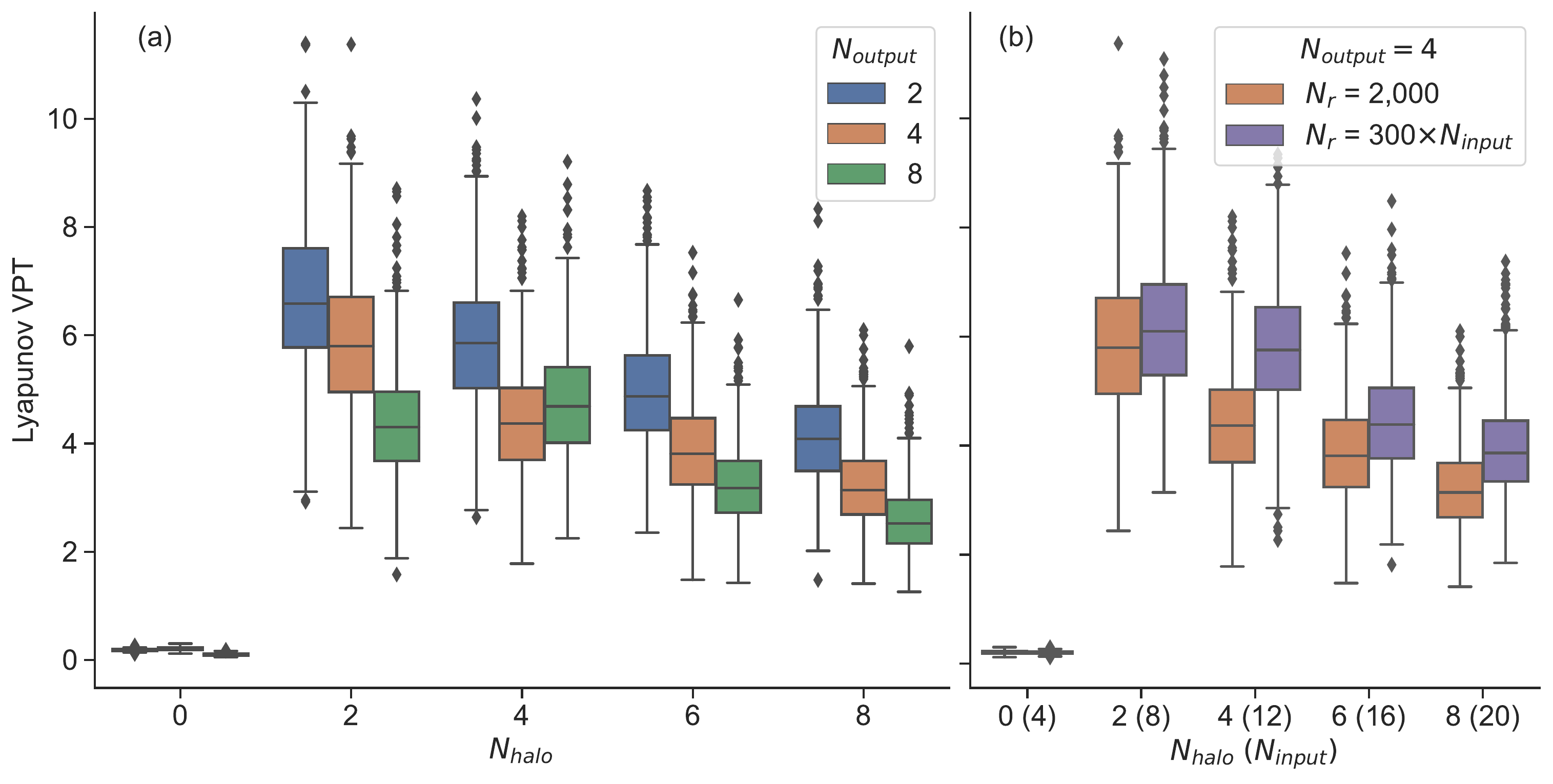}
\caption{VPT of localized RC model predictions for the 40D Lorenz96 system. 
    (a) Performance across various  $N_\text{output}$ (color) and $N_\text{halo}$ values (x-axis). The reservoir size is fixed at $N_r = 2,000$.
    (b) Performance for $N_\text{output}=4$, with the reservoir size fixed at 2,000 (orange) and reservoir size increasing with $N_\text{input}$ (purple).
    Each histogram is generated from 1,000 samples with parameters optimized for the given reservoir size, output dimension, and halo size. Parameter values are given in Tables \ref{tab:localrc_a}\&\ref{tab:localrc_b}.
    }
\label{fig:localrc}
\end{figure}

Figure \ref{fig:localrc}(a) shows the performance of localized RC models with varying output size,
$N_\text{output}$ and halo size, $N_\text{halo}$.
For output dimensions 2 and 4 (color), we see that the RC model shows its best performance when the halo size is 2. These RC models show no prediction skill with halos smaller than 2, and exhibit diminishing performance as the halo size increases beyond this point. These results suggest that for this particular system, the optimal configuration has a halo size of 2.
For the Lorenz96 system, this is unsurprising since we know that the time evolution for each individual node requires information of neighboring states up to 2 nodes away, equation (\ref{Lorenz-96}). For halo sizes smaller than 2, the known interactions between neighboring grid points are not represented. As the halo size increases, the RC model must effectively `learn' the length scale across which the underlying dynamical interactions occur.

The results suggests that the optimal halo size should be set to a ``minimum length scale'' relevant to the time evolution of the dynamical system. This value is obviously system dependent, and will require knowledge of how the training data are acquired. If the data arise from a numerical model, the halo size could be related to the numerical stencil used in time integration (e.g.\ finite difference or finite
element scheme). In the case of observed data, approximating the minimum length scale will require knowledge of the system dynamics.

As the output dimension for each local reservoir increases we see a similar effect as increasing the halo size: the prediction skill decreases.
Taken together, prediction skill is optimal when the reservoir ``stencil'' is as small as possible: each local RC model is trained with exactly the amount of
information it needs. This result is in some ways a restatement of \cite{pathak18} (their Figure 5b), who show that prediction performance improves as the number of local reservoirs increases with a fixed
system dimension. Using more local reservoirs requires more computational resources, and the benefits must be weighed against the increase in computational cost.

\subsection{Increasing the reservoir size with input dimension}
\label{ssec:fixed_ri_ratio}

The previous section showed that prediction skill decreases as the input dimension increases.
Here, we test the impact of increasing the reservoir size in order compensate for this reduction in performance. Figure \ref{fig:localrc}(b) shows the result of this test for $N_\text{output}=4$, comparing performance with a fixed reservoir size $N_r=2,000$ (orange) against models where $N_r=300N_\text{input}$. 
For $N_\text{halo}=4$, the median VPT increases by ~1.4 (32\%) as a result of increasing the reservoir size from 2,000 to 3,600 (80\%). For larger values of $N_\text{input}$, the payoff from increasing the reservoir size is even smaller. For example, $N_\text{halo}=6$, the median VPT increases by ~0.6 (15\%) but requires a reservoir size that is 4,800 (140\% larger). The results highlight what is shown in Section \ref{ssec:reservoir_dimension} (Figure \ref{fig: res_dim}). After a certain level of performance is obtained for a given problem size, increasing the reservoir size provides diminishing returns for prediction skill. 

\subsection{Input bias at scale}
\label{ssec:bias_at_scale}

\begin{figure}
\centering
\includegraphics[width=.6\textwidth]{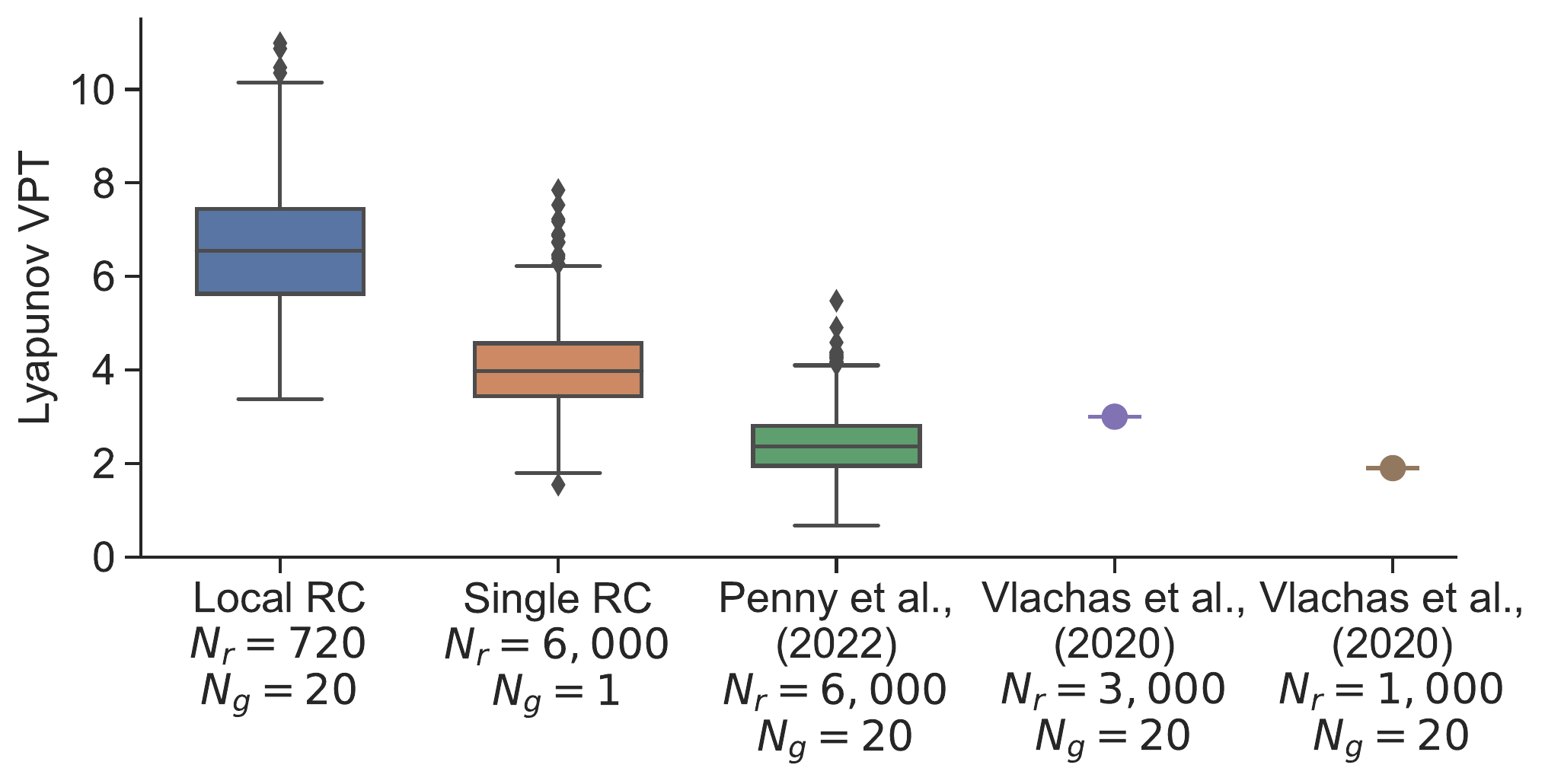}
\caption{VPT of various RC based predictions of the 40D Lorenz96 system.
    The Local RC (blue) has the configuration $(N_\text{output},N_\text{halo}, N_r)=(2,2,720)$.
    The Single RC (orange) uses $N_r=6,000$ (Figure \ref{fig:why_scale_up}).
    The green box plot shows results from the localized RC Model 3 in \cite{penny2021}, which uses $(N_\text{output}, N_\text{halo}, N_r) = (2, 4, 6000)$.
    The purple and tan dotted lines indicates the VPT from \cite{Vlachas20}, which use $(N_\text{output}, N_\text{halo}, N_r) = (2, 4, 3000)$ and $(N_\text{output}, N_\text{halo}, N_r) = (2, 4, 1000)$, respectively.
    We note that the VPT computed by \cite{Vlachas20} used a threshold of 0.5, so we estimate their VPT based on a threshold of 0.3 to match our results.
    Additionally, \cite{Vlachas20} compute VPT based on the average NRMSE evolution from 100 sample points, while we compute a histogram of VPT based on each NRMSE from 1,000 sample points.
    Both the Local RC and Single RC models use an optimized nonzero input bias $\sigma_b$, while the other RC designs use $\sigma_b=0$.
    Parameter values for the first three models are given in Table \ref{tab:performance_showdown}.
}
\label{fig:performance_showdown}
\end{figure}

Finally, we emphasize the importance of the bias term in the RC architecture, which
is accentuated in high dimensional systems.
Figure \ref{fig:performance_showdown} compares the performance of two RC models that use an optimized value for $\sigma_b$ (``Local RC'' and ``Single RC'') against two RC models that use $\sigma_b=0$ (from \cite{penny2021} and \cite{Vlachas20}).
The RC configuration from \cite{penny2021} uses $N_g=20$ local reservoirs, each with an output dimension of 2, halo size 4, and reservoir size of 6,000. Even with all other parameters optimized, the VPT is roughly half that of a single RC model with reservoir size 6,000 when an optimized (nonzero) bias term is used. When 20 local reservoirs are used with an optimized bias term and
$(N_\text{output}, N_\text{halo}) = (2, 2)$, the VPT is approximately tripled.

Penny et al. (2021) \cite{penny2021} found it necessary to increase the local reservoir dimension to 6,000 in order to attain reasonable prediction skill.
The results in this section show that this was essentially a brute force solution to overcome an inactive bias term. This is an important takeaway for forecasting high dimensional systems, which may have such a large number of localized RC models that increasing the reservoir size to overcome this inactive bias term would become prohibitive. 


\section{Conclusion and Discussion}
Reservoir computing is a powerful machine learning method that can be used to successfully predict chaotic time series data. While not as flexible as other RNN methods, the RC has a number of properties that make it a good method of choice for these kinds of tasks. The ability to set the macro-scale properties of the network and the quick training of the micro-scale parameter through linear regression couple together to allow the RC to be trained and deployed quickly and easily. Additionally, the RC has been shown not only to give good predictions but also to react like a physical model to perturbations in the system state (shown as the reproduction of the LEs) \cite{Lu18, platt21, penny2021}. This property is crucial for applications in numerical weather prediction and other fields where it is important to produce both a forecast as well as an uncertainty estimate of that forecast. RCs can also be scaled up to large systems while being accelerated on dedicated hardware for vast speedups \cite{canaday18}, thus making them an exciting candidate for the simulation of high dimensional systems.

We present in this paper an exploration of the options needed to for RCs to be trained successfully for chaotic time series forecasting. Almost all the results presented here involve the reoptimization of parameters through the outlined Bayesian optimization procedure for every experiment. The complex effects of the macro-scale parameters on the predictability of the RC and the correlations between these parameters preclude drawing many conclusions from simply varying each one of the parameters individually. When the data, RC training method, or one parameter is changed then a full re-optimization is necessary to compensate. This is one reason why the results presented here differ from those presented in previous literature.

The bias term in the RC equation is found to be criticaly important Fig.(\ref{fig: input_bias}), for all but the L63 system, and  we emphasize that this is generally neglected in the current literature. Indeed predictions were practically impossible for many of the models without the addition of this term. We additionally show that the form of the readout has little impact on the skillful forecast time, at least for the given dynamical models. While this may be surprising given that the inclusion of an $r^2$ term in the readout has become a standard practice \cite{Lu18, pathak18, Vlachas20}, we note that without the inclusion of bias, the nonlinear readout does have some positive impact (results not shown).

Our further exploration included showing the deleterious effects of a standard ML normalization procedure where each dimension of the input is recentered and rescaled separately \cite{Goodfellow16}. This procedure is important in deep learning where the stability of the optimization procedure using backpropagation can be compromised. Centering all of the data around 0, however, destroys the important relationships between the variables, leading to a very poor predictive capacity for the RC. The RC is not trained through backpropagation, therefore it is advisable to avoid this kind of normalization when attempting to predict chaotic time series.

We also examine the effect of the reservoir dimension $N$, the amount of training data as well as the time step of the data on the performance of an RC. A general ``law of diminishing returns'' is exhibited in regards to RC size and the amount of data. This effect can be partially mitigated by re-optimizing the RC using a larger reservoir size instead of scaling up directly. We suggest that more effort should be dedicated to increasing the robustness of the RC to noise, since in real applications this will be a limitation on the usefulness of the method.

Finally, we show how to scale up the method to larger systems through localization, examining the effect of halo size and output dimension. We present results on the 40 dimensional Lorenz system. Comparing to state of the art results for the RC \cite{Vlachas20, penny2021}, the reported architecture in this paper increases the VPT while decreasing the computational cost. Vlachas et al. (2020) \cite{Vlachas20} noted ``that RC...have slightly lower VPT than GRU and LSTM but require significantly lower training times.'' Our results now give a clear advantage of RC over the LSTM and GRU in pure predictive capacity as well as training time; additionally the RC has the advantage of reproducing the LEs of the input data, enabling the reproduction of the climatalogical attractor \cite{Lu18}. We hope that this work, by collecting a number of different characteristic models together, gives a clear set of standards to compare as a benchmark when developing new RC techniques.

\section{Acknowledgements}

J.A. Platt, S.G. Penny, and H.D.I. Abarbanel acknowledge support from the Office of Naval Research (ONR) grants N00014-19-1-2522 and N00014-20-1-2580. S.G. Penny and T.A. Smith acknowledge support from NOAA grant NA20OAR4600277. T.-C. Chen is supported by the NOAA Cooperative Agreement with CIRES, NA17OAR4320101. S.G. Penny and J.A. Platt jointly proposed and guided the research study. J.A. Platt conducted the majority of the investigation. T.A. Smith conducted the study of localization and scaling to larger systems. Coauthors Penny, Platt, Smith, and Chen all contributed to the development of the software used to conduct the experiments in this study.

\printbibliography

\appendix

\section{Lyapunov Exponents} \label{app: LE}
Given an N-dimensional dynamical system \footnote{$f_x(\x(t), t)$ defines a dynamical system with external forcing.  Formally, the LEs calculated for this system are called conditional Lyapunov exponents (CLE). If $f_x(\x(t))$ does not depend explicitly on time then it is an autonomous system and we are just calculating LEs.}
\begin{equation}
    \dot \x(t) = f_x(\x(t), t) \label{eq: dynamical sys}
\end{equation}
with $\x$ the state, $f$ the dynamics and $\cdot$ denoting $\dv{t},$ the spectrum of LEs $\mqty[\lambda_1, \lambda_2,...,\lambda_N]$ can be computed as follows. First we find the linearization of the system. We derive the variational equations, which describe the evolution of the tangent vectors, by taking $\pdv{x_j(t')}$ on both sides of eq \eqref{eq: dynamical sys}. Applying the chain rule we find
\begin{equation*}
    \dv{t} \pdv{x_i(t)}{x_j(t')} = \pdv{f_i(x(t), t)}{x_k(t)} \pdv{x_k(t)}{x_j(t')}
\end{equation*}
which we rewrite as
\begin{equation}
\dot \bphi(t', t) = \J(x(t))\cdot \bphi(t', t) \label{eq: variational}
\end{equation}
with $\bphi(t', t) = \pdv{x_i(t)}{x_j(t')}$ being the $N \times N$ variational matrix and $\J(x(t)) = \J(t) = \pdv{f_i(x(t), t)}{x_k(t)}$ being the Jacobian.  The general solution to Eq.\eqref{eq: variational} involves the matrix exponential
\begin{equation*}
\bphi(t', t) = \mathcal{T}_+ \exp{\int_{t}^{t'}\J(x(t)) dt} \bphi(t,t)
\end{equation*}
with $\bphi(t, t) = \mathbb{I}$ 
and $\mathcal{T}_+$ being the \textit{time ordering operator} \cite{db64} for two operators (here matrices) $\A(t)$ and $\B(t')$ that are not commutative, 
\begin{equation*}
    \mathcal{T}_+ \A(t) \B(t') = \begin{cases}
                                    \A(t)\B(t') & \mbox{if } t > t' \\
                                    \B(t')\A(t) & \mbox{if } t < t' \\
                              \end{cases}.
\end{equation*}
$\mathcal{T}_+$ is important because the Jacobian matrices are not commutative \ie, $\mqty[\J(t),\J(t')] \ne 0.$ 

The variational equation is much easier to work with in discrete time $t = t_0 + n \Delta t;\;t' = t_0 + n' \Delta t$
where it appears as 
\begin{equation*}
    \bphi(n',n) = \DF(n) \bphi(n'-1,n),
\end{equation*}
and $\bphi(n,n) = \mathbb{I}$. 

Then the solution for $\bphi(n+L,n)$ is the product of Jacobians
\begin{eqnarray}
\bphi(n+L,n) &&= \DF(n + L) \cdot \bphi(n+L-1,n) \nonumber \\
&&= \DF(n+L) \cdot \DF(n+L-1) \cdot \bphi(n+L-2) \nonumber \\
&&= \DF(n+L) \cdot \DF(n+L-1) \cdot \ldots \DF(n).
\end{eqnarray}

The matrix $\bphi(t', t)$ describes how small perturbations to a state $x(t)$ propagate to the state at $x(t')$. Given equations \eqref{eq: dynamical sys} and \eqref{eq: variational} one can solve them concurrently to find Oseledec's matrix $\mathbf{\Phi} = \bphi(t',t)\bphi(t',t)^T.$  

The eigenvalues of the $\log$ of $\mathbf{\Phi}(t',t)$ for large times
\begin{equation}
    \lim_{t' \to \infty} \frac{1}{2 t'} \log{\bphi(t', t)\bphi(t', t)^T} \label{eq: Lambda}
\end{equation}
are the \textbf{global} Lyapunov Exponents; the $N$ eigenvalues are by definition real and the eigenvectors orthogonal since $\mathbf{\Phi}$ is a symmetric $N \times N$ matrix. We order the LEs $\lambda_1 > \lambda_2 > \ldots > \lambda_N$. The matrix $\mathbf{\Phi}$ is ill-conditioned, so accurately evaluating all of its eigenvalues requires a stable algorithm, for example as provided by ~\cite{Eckmann85, Abarbanel96_book}.

Oseledec's multiplicative ergodic theorem~\cite{Oseledec68} states that all $N$ LEs of a dynamical system (a) exist, (b) are independent of the initial starting point $x(t)$, and (c) are invariant under smooth coordinate transformations \footnote{The statement of invariance under coordinate transformation is important when attempting to recover the LEs from a time delay embedding.}.  The finite time Lyapunov exponents (FTLE) \cite{Abarbanel92}, Eq.(\ref{eq: Lambda}) evaluated for finite $t'$, are not independent of $x(t)$ nor are they invariant under a smooth coordinate transformation.  Furthermore, for a continuous time dynamical system one of the LEs must be 0, and $\sum_{i=1}^N \lambda_i \le 0.$  In Hamiltonian systems the spectrum of LEs is symmetric around 0 and thus $\sum_{i=1}^N \lambda_i = 0,$ a natural statement of phase space volume conservation guaranteed by Liouville's theorem.

These definitions carry over to the forced dynamics with vector field $f_r(\r(t),\u(t))$, which we find in driven RNNs such as the driven RC Eq.\eqref{eq: mapping_form} considered here. They are then called conditional Lyapunov exponents (CLEs) as they are conditional on the driving forces $\u(t)$~\cite{Pecora90}.


\section{Malkus Water Wheel} \label{app: malkus}
For physical applications, determining the correct normalization scheme can be more difficult when the state variables have different units \eg, temperature, pressure, velocity,\ldots. All of the source models that we used in this study were nondimensionalized. In practice we may instead have data measured in quantities with different units. There are two solutions that we propose:
\begin{enumerate}
    \item Nondimensionalize the state variables using physical parameters of the system. This does not require knowledge of the full equations of motion, only basic knowledge of the physics involved.
    \item Allow the RC to learn the correct scaling laws by setting a separate $\sigma$ and $\sigma_b$ for each unit type. For example, one might have a $\sigma_{\rm Temperature}$ and a $\sigma_{\rm Pressure}$. Note that this replaces the scalar $\sigma$ with a vector $\vb \sigma$ that has the structure $\vb \sigma = \mqty[\sigma_1,\ldots, \sigma_1,\ldots, \sigma_p]$ for $p$ different units for the state variables. 
\end{enumerate}

As an illustrative example, we consider the Malkus water wheel \cite{strogatz00, Matson07, Illing12}. The chaotic water wheel is a physical model of the L63 equations and thus provides a mixed-units example with which to apply the proposed normalization schemes. Following the notation of \cite{Matson07}, the three state variables are $\omega$, $y$ and $z$ where $\omega$ is the angular velocity of the water wheel $\dv{\theta}{t}$ for the angle $\theta$ in the plane of the wheel, and $y/z$ gives the position of the center of mass (COM) also in the plane of the wheel. The equations of motion are
\begin{align*}
    &\dot \omega = a y - f \omega\\
    &\dot y = \omega z - \lambda y\\
    &\dot z = -\omega y + \lambda (R - z)
\end{align*}
where $a$ is the angular acceleration due to gravity per horizontal displacement of the COM, $f^{-1}$ is the time constant for the axle friction and input water drag, $\lambda$ is the leakage rate of the water and $R$ is the radius of the wheel. We see that $\lambda^{-1}$ also gives the relaxation time constant for the COM.  The maximal LE when $R = 1\ m$, $a = 1\ (m\ s)^{-1}$, $f=0.4$, $\lambda=0.1\ s^{-1}$ is $0.053$ giving a time constant of $\sim 19$ model time units.

To nondimensionalize the system we see that there are two different units for the state variables. $\omega$ has units $1/\rm{time} \equiv 1/T$ while $y$ and $z$ have units $\rm{length} \equiv \ell$. Therefore, if we choose a time scale $T$ and length scale $\ell$, we can nondimensionalize by taking $\omega_\star = \omega/T$, $y_\star = \ell y$ and $z_\star = \ell z$. Using knowledge of the physics but without knowing the equations of motion we can write a list of parameters on which the equations may depend---see table \ref{tab: malkus}. Again, no a priori knowledge of the equations of motion is needed.
\begin{table}[!htpb]
    \centering
    \begin{tabular}{c|c|c}
         $\alpha$ & axel friction & $kg\ m/s^2$\\
         $I$ & moment of inertia & $kg\ m^2$\\
         $\lambda$ & leak rate & $1/s$\\
         $R$ & radius & $m$\\
         $g \sin \phi$ & gravity at angle $\phi$& $m/s^2$\\
         $M$ & mass & $kg$\\
    \end{tabular}
    \caption{Deduced parameters of the equations of motion for the Malkus water wheel.}
    \label{tab: malkus}
\end{table}

We can construct $T$ and $\ell$ using this set of candidate parameters. An obvious choice for $\ell$ is $R$, since both have the same units. For $T$ we could use $T = 1/\lambda$ or $T = \sqrt{\frac{RMg\sin \phi}{I}}$. There is no wrong answer as long as all the parameters with the same units are scaled in the same way and one can measure the parameters needed.

\begin{figure}[!htpb]
    \centering
    \includegraphics[width=0.45\textwidth]{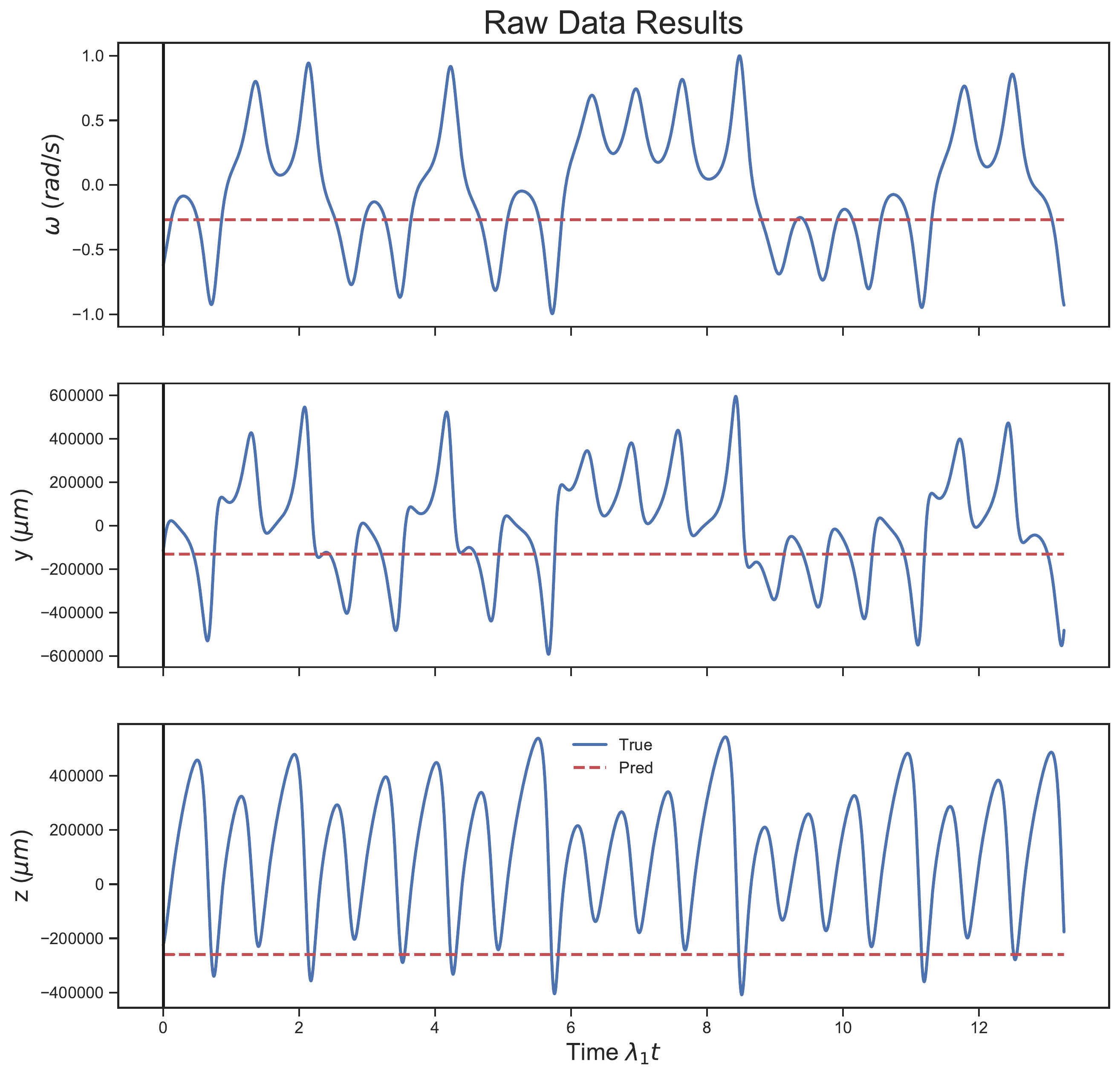}
    \includegraphics[width=0.45\textwidth]{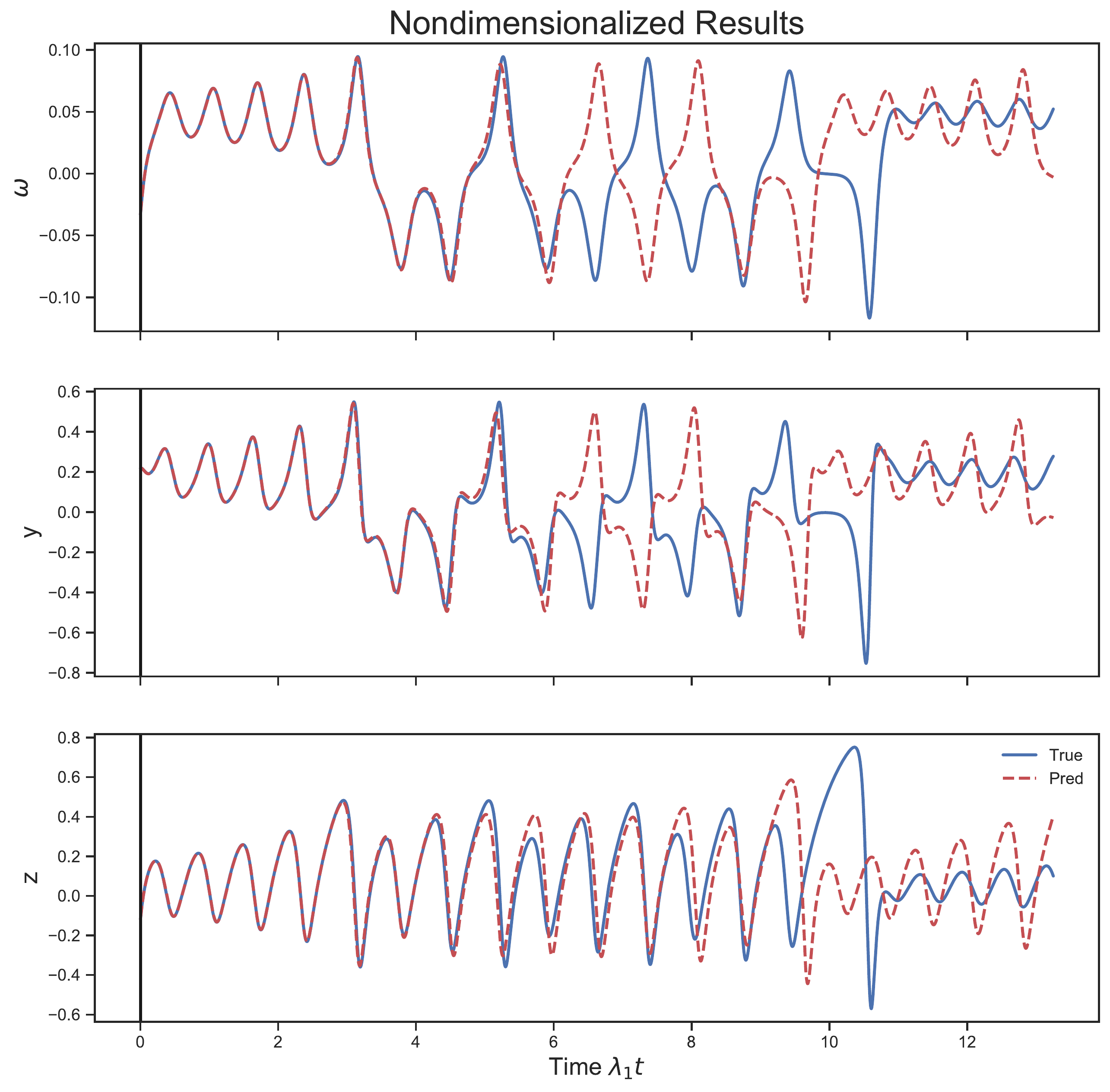}
    \caption{\textbf{(left)} Our hypothetical sensor measures the COM in $\mu m$, leading to an RC that utterly fails.  \textbf{(right)} By normalizing the measurements correctly we find that the RC can now predict the Malkus water wheel with reasonable accuracy.}
    \label{fig: malkus-micron}
\end{figure}

As a (somewhat extreme) example, let us say the distance measurements from our sensor are in $\mu m$, and $\omega$ is measured in $\rm rad/s$. In this case the RC is not able to predict the system at all Fig.(\ref{fig: malkus-micron}). This is not really surprising considering the wildly different scales between $\omega$ and the $x$ and $y$ variables. We correct this by scaling $\omega$ by $\lambda$ and $x$ and $y$ by $R$. This leads to reasonably accurate forecasts. Letting the optimization determine the correct scaling between the variables also leads to accurate forecasts.

\section{Details of Datasets}
\subsubsection{Rossler}
The Rossler system \cite{rossler76} was proposed as a simplified form of the L63 system with a single second order nonlinearity and only one lobe of the strange attractor. The equations are
\begin{align*}
&dx = -(y + z)\\
&dy =  x + 0.2  y\\
&dz =  0.2 + z (x - 5.7).\\
\end{align*}
The LEs  are $\mqty[ 0.06, 0, -4.9]$.
\begin{figure}[!htpb]
    \centering
    \includegraphics[width = 0.5\textwidth]{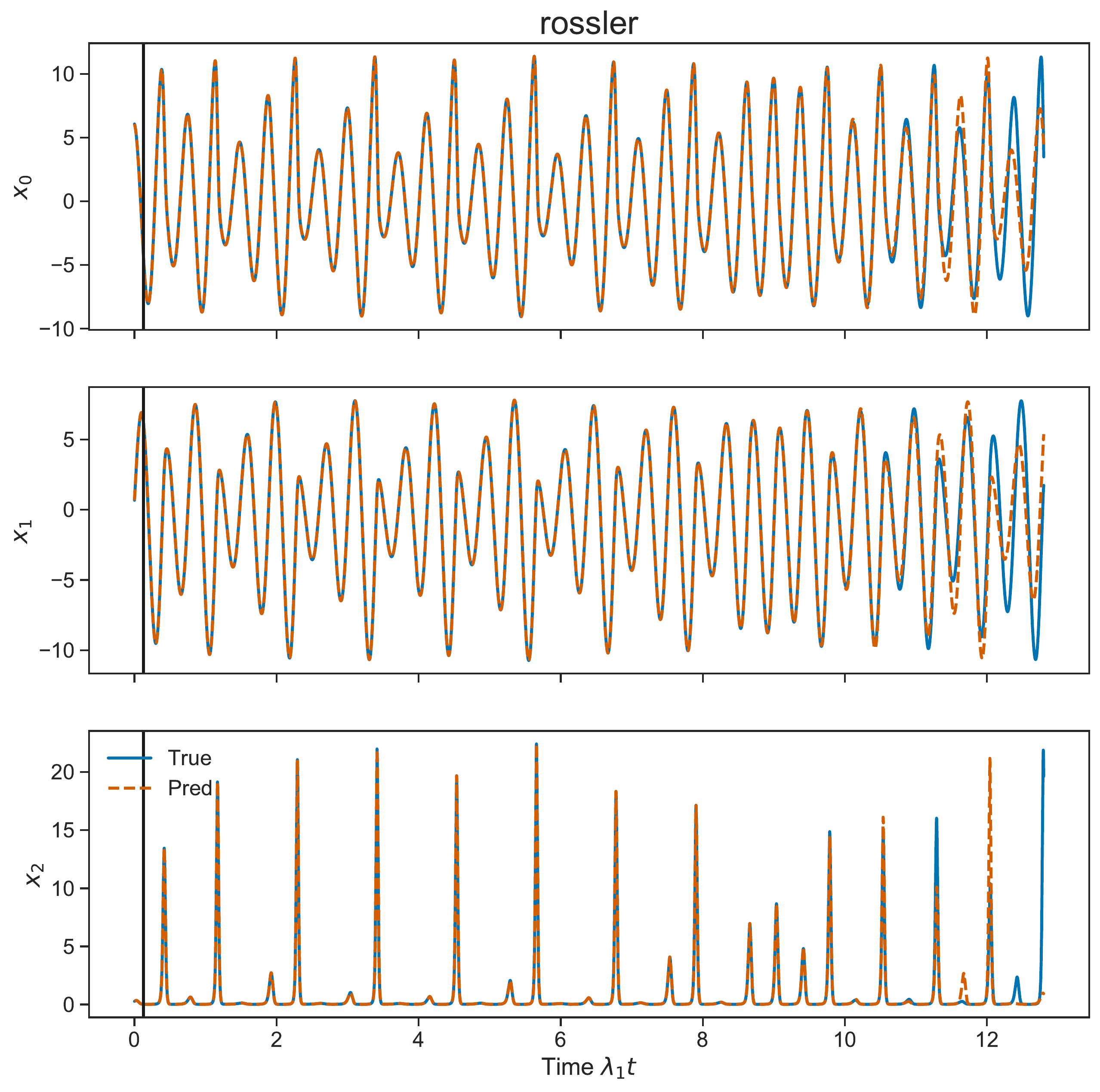}
    \caption{Rossler Attractor}
    \label{fig: prediction_rossler}
\end{figure}
\subsubsection{Colpitts}
The Colpitts Oscillator \cite{colpitts18} is a three dimensional nonlinear dynamical system describing chaos in a nonlinear circuit. The equations of the system are given
\begin{eqnarray*}
    \frac{dx(t)}{dt} &=& \alpha y(t) \nonumber \\
    \frac{dy(t)}{dt} &=& -\gamma (x(t)+z(t))-q y(t)  \nonumber \\
    \frac{dz(t)}{dt} &=& \eta (y(t) +1 -\exp(-x(t))) 
\end{eqnarray*}
with $\alpha = 5$, $\gamma = 0.0797$, $q = 0.6898$ and $\eta = 6.2723.$ For $\alpha < 5$ this circuit has limit cycle oscillations. 

The Lyapunov exponents are $\{\lambda_1, \lambda_2,\lambda_3\} = \mqty[0.09,& 0,&  -0.8]$ calculated via the QR decomposition algorithm given by Eckmann and Ruelle~\cite{Eckmann85}.
\begin{figure}[!htpb]
    \centering
    \includegraphics[width = 0.5\textwidth]{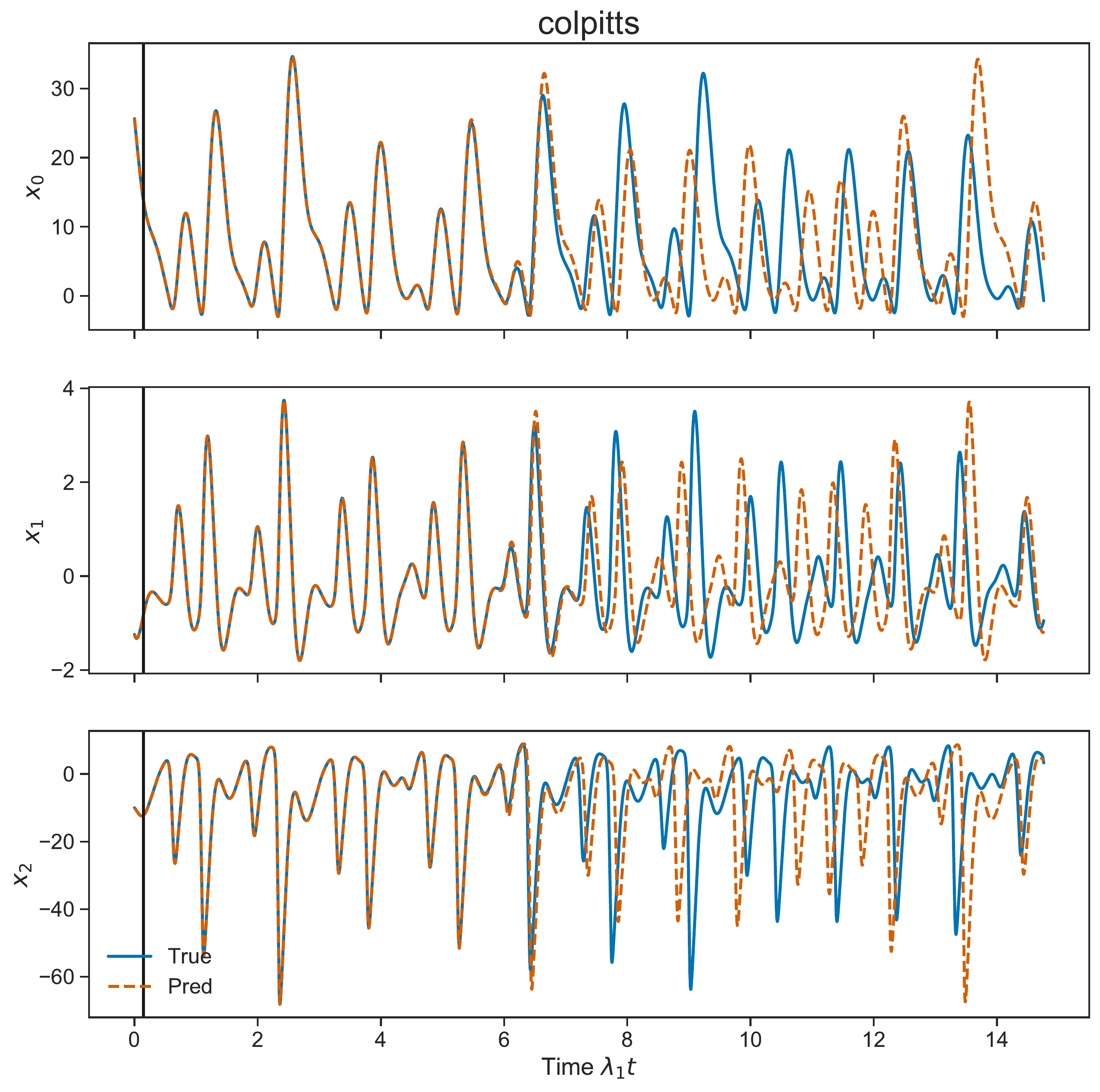}
    \caption{Colpitt's Oscillator}
    \label{fig: prediction_colpitts}
\end{figure}
\subsubsection{L63}
The Lorenz-63~\cite{Lorenz63} equations form a deterministic nonlinear dynamical system that exhibits chaos for certain ranges of parameters. It was originally found as a three dimensional, reduced, approximation to the partial differential equations for the heating of the lower atmosphere of the Earth by solar radiation. The dynamical equations of motion are
\begin{eqnarray*}
&& \frac{dx(t)}{dt} = \sigma[y(t) - x(t)] \nonumber \\
&&\frac{dy(t)}{dt}     = x(t)[\rho - z(t)] - y(t) \nonumber \\
&&\frac{dz(t)}{dt} = x(t)y(t) - \beta z(t)\\
\label{lor63}
\end{eqnarray*}
with time independent parameters $\sigma = 10, \rho = 28, \beta = 8/3$.

The Lyapunov exponents are $\{\lambda_1, \lambda_2,\lambda_3\} = \mqty[0.9, &  0, & -14.7]$ calculated using the QR decomposition approach given by Eckmann and Ruelle~\cite{Eckmann85}.

\begin{figure}[!htpb]
    \centering
    \includegraphics[width = 0.5\textwidth]{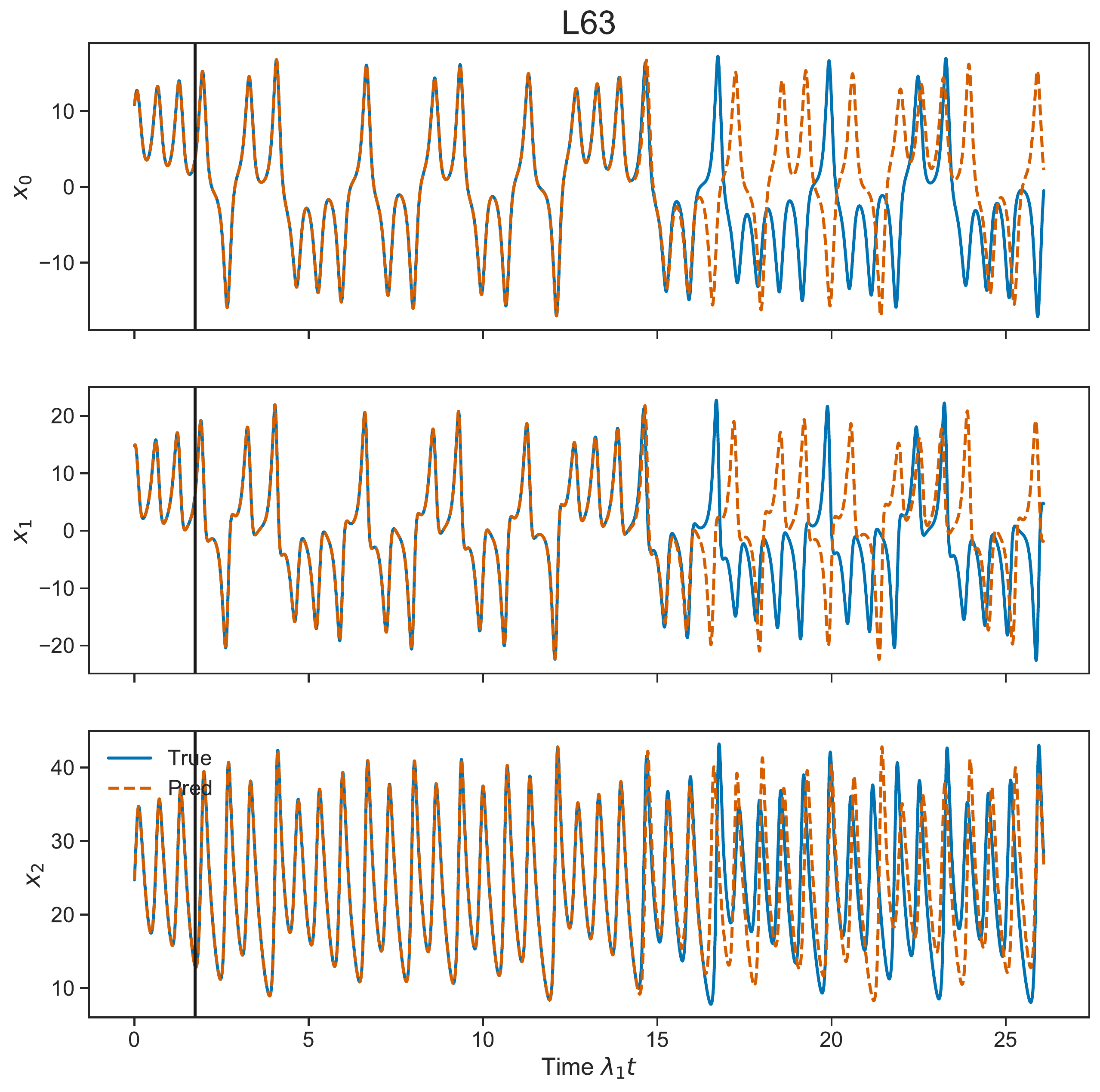}
    \caption{Lorenz 1963}
    \label{fig: prediction_L63}
\end{figure}

\subsubsection{L96}
These dynamical equations were introduced by~\cite{Lorenz96}:
\begin{equation*}
\frac{dx_a(t)}{dt} = x_{a-1}(t)(x_{a+1}(t) - x_{a-2}(t)) - x_a(t) + f
\label{Lorenz-96}
\end{equation*}
and $a=1,2,...,D$; $x_{-1}(t) = x_{D-1}(t)$; $x_0(t) = x_D(t)$; $x_{D+1}(t) = x_1(t)$. $f$ is a fixed parameter that we take to be  8.0 where the solutions to these dynamical equations are chaotic~\cite{kostuk}. The equations for the states $x_a(t);\; a = 1, 2, ..., D$ are meant to describe points on a periodic spatial lattice. We use $D = 5$, $D = 10$ and $D = 40$.

The Lyapunov exponents are $\{\lambda_1, \ldots,\lambda_5\} = \mqty[0.4,  0, -0.5, -1.3, -3.5]$ and $$\{\lambda_1, \ldots,\lambda_{10}\} = \mqty[1.1,  0.7,  0.1,  0, -0.4, -0.8, -1.3, -1.9, -2.7, -4.5]$$ calculated using the QR decomposition approach given by Eckmann and Ruelle~\cite{Eckmann85}.

\begin{figure}[!htpb]
    \centering
    \includegraphics[width = 0.5\textwidth]{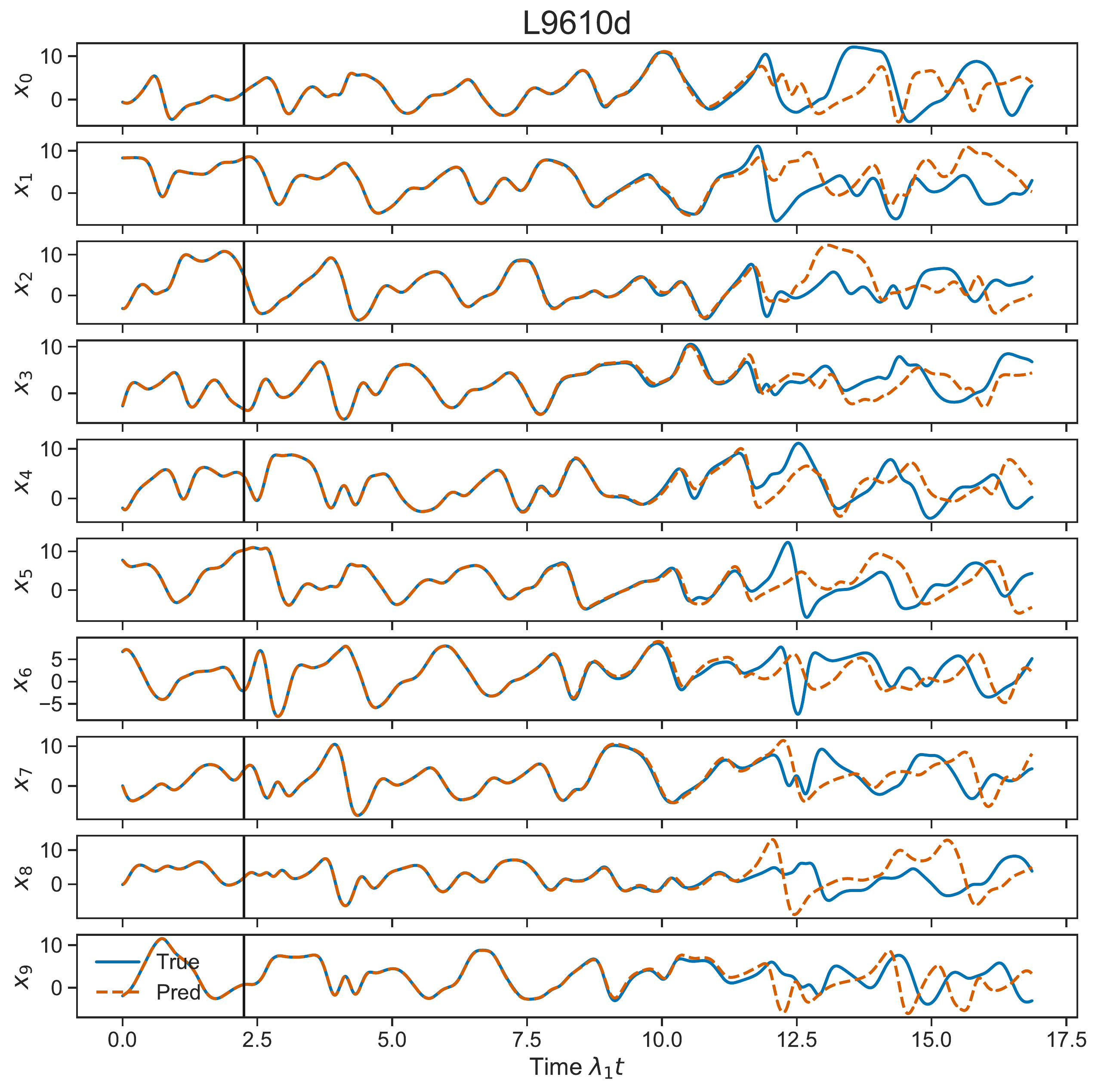}
    \caption{Lorenz 1996 10 dimensions}
    \label{fig: prediction_L9610d}
\end{figure}

\subsubsection{CL63}
The climate Lorenz 63 system\cite{pena04} consist of three L63 systems coupled together with multiple time scales. The three layers are an extratropical atmosphere, a tropical atmosphere, and a tropical ocean system. The presence of multiple time scales often creates a challenge for RC models.
\begin{align*}
        &\text{Extratropical atmosphere system}\\
        &\dv{x_e}{t} = \sigma (y_e - x_e) - \kappa_e(S x_t + k_1)\\
        &\dv{y_e}{t} = \rho x_e - y_e - x_e z_e + \kappa_e (S y_t + k_1)\\
        &\dv{z_e}{t} = x_e y_e - \beta z_e\\
        &\text{Tropical atmosphere system}\\
        &\dv{x_t}{t} = \sigma (y_t - x_t) - \kappa (S x_o + k_2) - \kappa_e (S x_e + k_1)\\
        &\dv{y_t}{t} = \rho x_t - y_t - x_t z_t + \kappa (S y_o + k_2) + \kappa_e (S y_e + k_1)\\
        &\dv{z_t}{t} = x_t y_t - \beta z_t + \kappa_z z_o\\
        &\text{Tropical ocean system} \\
        &\dv{x_o}{t} = \tau \sigma (y_o - x_o) - \kappa (x_t + k_2)\\
        &\dv{y_o}{t} = \tau \rho x_o - \tau y_o - \tau S x_o z_o + \kappa (y_t + k_2)\\
        &\dv{z_o}{t} = \tau S x_o y_o - \tau \beta z_o - \kappa_z z_t\\
\end{align*}
$\sigma=10, \rho=28, \beta=8/3., S=1, k_1=10, k_2=-11, \tau=0.1, \kappa=1, \kappa_e=0.08, \kappa_z=1$
$LEs = \mqty[0.9,  0.4,  0, -0.1, -0.6, -0.8, -1.6, -11.7, -14]$

\begin{figure}[!htpb]
    \centering
    \includegraphics[width = 0.5\textwidth]{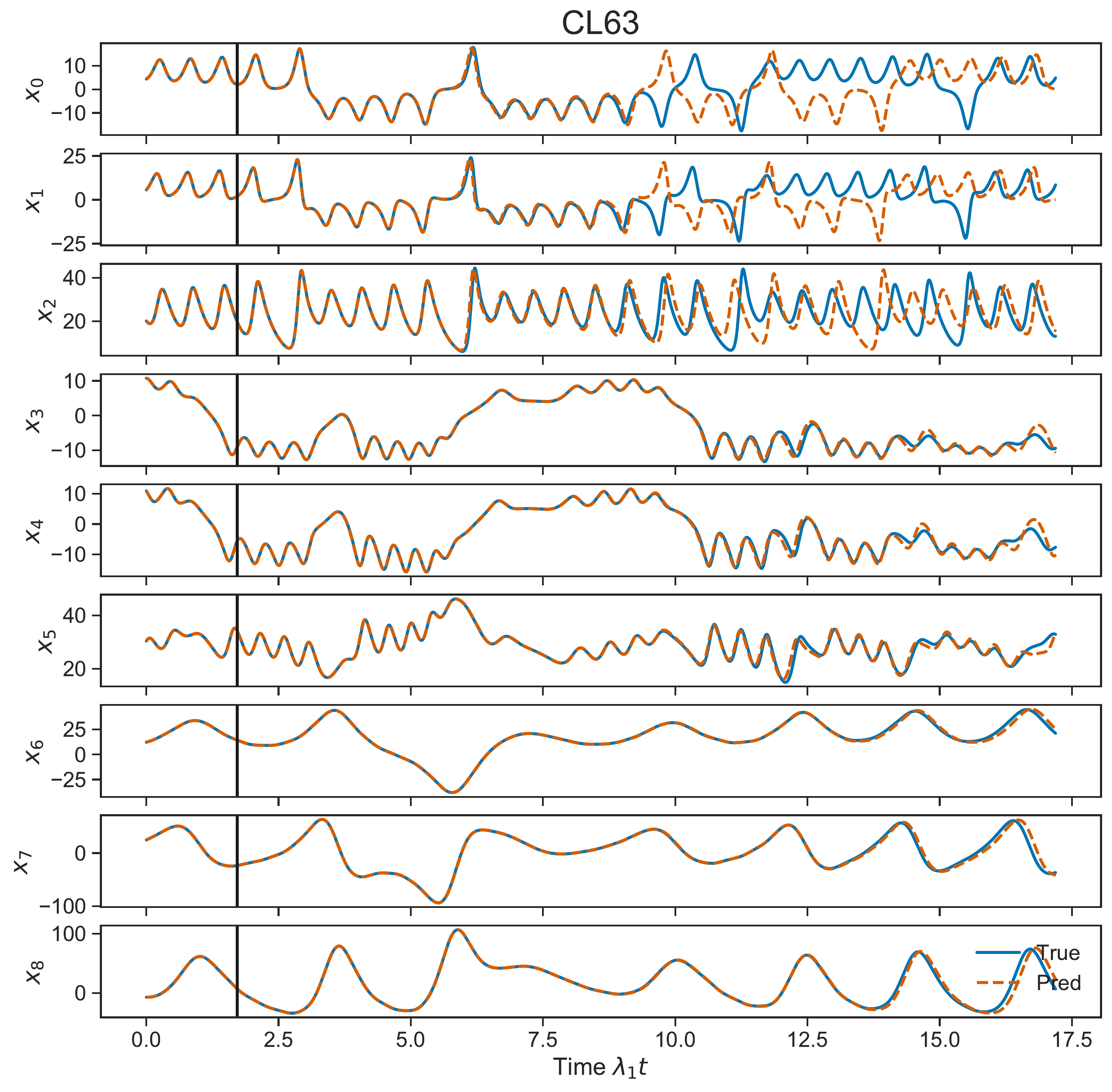}
    \caption{Climate Lorenz 63 model}
    \label{fig: prediction_CL63}
\end{figure}

\section{Parameters}
\begin{table}[!htpb]
    \centering
    \begin{tabular}{lrrrrrrlr}
    \toprule
     dataset & $\sigma$ & $\alpha$ &   $\rho_{SR}$ & $\beta$ & N & $\sigma_b$ &   readout & $\rho_A$ \\
    \midrule
        CL63 & 0.007 &  1.00 & 0.01 & 2.201375e-08 & 2000 &    0.66 &    linear &  0.98 \\
         L63 & 0.084 &  0.60 & 0.80 & 8.493901e-08 & 2000 &    1.60 & linear &  0.98 \\
      L9610d & 0.005 &  0.72 & 0.21 & 7.640822e-09 & 2000 &    1.47 &    linear &  0.98 \\
       L96-5D & 0.060 &  0.70 & 0.58 & 6.332524e-09 & 2000 &    1.59 &    linear &  0.98 \\
    colpitts & 0.100 &  1.00 & 1.20 & 1.000000e-08 & 2000 &    2.00 &    linear &  0.98 \\
     rossler & 0.066 &  0.47 & 0.50 & 2.101845e-09 & 2000 &    1.23 & linear &  0.98 \\
    \bottomrule
    \end{tabular}
    \caption{Fig \ref{fig: best}}
    \label{tab: best}
\end{table}

\begin{table}[!htpb]
    \centering
\begin{tabular}{lrrrrrrlr}
\toprule
 dataset & $\sigma$ & $\alpha$ &   $\rho_{SR}$ &         $\beta$ &    N & $\sigma_b$ &   readout & $\rho_A$ \\
\midrule
    CL63 & 0.002 &  0.84 & 0.53 & 1.033811e-07 &  250 &    1.31 &    linear &  0.98 \\
    CL63 & 0.007 &  1.00 & 0.01 & 2.201375e-08 & 2000 &    0.66 &    biased &  0.98 \\
     L63 & 0.024 &  0.96 & 0.64 & 1.339452e-08 &  250 &    1.63 &    biased &  0.98 \\
     L63 & 0.084 &  0.60 & 0.80 & 8.493901e-08 & 2000 &    1.60 & quadratic &  0.98 \\
  L9610d & 0.005 &  0.51 & 0.68 & 2.844557e-10 &  250 &    1.37 &    biased &  0.98 \\
  L9610d & 0.005 &  0.72 & 0.21 & 7.640822e-09 & 2000 &    1.47 &    biased &  0.98 \\
   L96-5D & 0.022 &  0.85 & 0.34 & 5.756201e-08 &  250 &    0.41 &    biased &  0.98 \\
   L96-5D & 0.060 &  0.70 & 0.58 & 6.332524e-09 & 2000 &    1.59 &    biased &  0.98 \\
colpitts & 0.100 &  1.00 & 0.01 & 3.548470e-09 &  250 &    0.00 &    biased &  0.98 \\
colpitts & 0.100 &  1.00 & 1.20 & 1.000000e-08 & 2000 &    2.00 &    linear &  0.98 \\
 rossler & 0.100 &  1.00 & 1.20 & 1.000000e-08 &  250 &    2.00 & quadratic &  0.98 \\
 rossler & 0.066 &  0.47 & 0.50 & 2.101845e-09 & 2000 &    1.23 & quadratic &  0.98 \\
\bottomrule
\end{tabular}
    \caption{Fig \ref{fig: res_dim}}
    \label{tab: res_dim}
\end{table}

\begin{table}[!htpb]
\centering
\begin{tabular}{lrrrrrrlr}
\toprule
 dataset & $\sigma$ & $\alpha$ &   $\rho_{SR}$ & $\beta$ &      N & $\sigma_b$ & readout & $\rho_A$ \\
\midrule
    CL63 & 0.004 &  0.98 & 0.66 & 5.216114e-10 & 1200 &    1.57 &  linear &  0.98 \\
    CL63 & 0.014 &  0.88 & 0.01 & 1.572121e-05 & 1200 &    0.00 &  linear &  0.98 \\
     L63 & 0.011 &  0.49 & 0.59 & 1.115693e-10 & 1200 &    0.28 &  linear &  0.98 \\
     L63 & 0.014 &  0.89 & 1.02 & 2.390891e-08 & 1200 &    0.00 &  linear &  0.98 \\
  L9610d & 0.017 &  0.95 & 0.10 & 1.443850e-09 & 1200 &    1.59 &  linear &  0.98 \\
  L9610d & 0.001 &  0.41 & 0.92 & 4.553807e-03 & 1200 &    0.00 &  linear &  0.98 \\
   L96-5D & 0.100 &  1.00 & 0.05 & 1.012426e-10 & 1200 &    0.55 &  linear &  0.98 \\
   L96-5D & 0.096 &  0.80 & 0.40 & 4.087443e-08 & 1200 &    0.00 &  linear &  0.98 \\
colpitts & 0.100 &  0.55 & 1.87 & 2.836222e-09 & 1200 &    2.85 &  linear &  0.98 \\
colpitts & 0.022 &  0.91 & 1.11 & 3.177257e-07 & 1200 &    0.00 &  linear &  0.98 \\
 rossler & 0.042 &  0.89 & 0.79 & 1.006571e-08 & 1200 &    0.71 &  linear &  0.98 \\
 rossler & 0.028 &  0.48 & 0.81 & 1.000000e-08 & 1200 &    0.00 &  linear &  0.98 \\
\bottomrule
\end{tabular}
\caption{Fig \ref{fig: input_bias}}
\label{tab: input_bias}
\end{table}

\begin{table}[!htpb]
    \centering
    \begin{tabular}{lrrrrrrlrr}
    \toprule
    dataset & $\sigma$ & $\alpha$ &   $\rho_{SR}$ & $\beta$ &    N & $\sigma_b$ & readout & $\rho_A$ & Train Data \\
    \midrule
     L9610d & 0.001 &  0.68 & 1.01 & 2.202428e-02 & 1000 &    1.40 &  linear &  0.98 &        100 \\
     L9610d & 0.003 &  0.85 & 0.43 & 1.043649e-09 & 1000 &    1.04 &  linear &  0.98 &       1000 \\
     L9610d & 0.006 &  1.00 & 0.01 & 1.769733e-10 & 1000 &    1.15 &  linear &  0.98 &      10000 \\
     L9610d & 0.008 &  0.22 & 0.02 & 9.382122e-08 & 1000 &    1.33 &  linear &  0.98 &     100000 \\
      L96-5D & 0.012 &  0.96 & 1.26 & 1.236301e-09 & 1000 &    3.00 &  linear &  0.98 &        100 \\
      L96-5D & 0.023 &  0.83 & 0.31 & 2.517159e-09 & 1000 &    1.23 &  linear &  0.98 &       1000 \\
      L96-5D & 0.039 &  0.69 & 0.85 & 1.349428e-08 & 1000 &    1.08 &  linear &  0.98 &      10000 \\
      L96-5D & 0.066 &  0.60 & 0.66 & 1.319216e-09 & 1000 &    1.04 &  linear &  0.98 &     100000 \\
    \bottomrule
    \end{tabular}
    \caption{Fig. \ref{fig: reg_training}}
    \label{tab: train data}
\end{table}

\begin{table}[!htpb]
    \centering
    \small
    \begin{tabular}{lrrrrrrlrl}
    \toprule
     dataset &  $\sigma$ & $\alpha$ &   $\rho_{SR}$ &         $\beta$ &      N & $\sigma_b$ &   readout & $\rho_A$ & normalization \\
    \midrule
        CL63 &  0.007 &  0.28 & 0.40 & 6.630623e-03 & 1512.0 &    1.44 & quadratic &  0.98 &      Scheme 1 \\
        CL63 &  1.566 &  1.00 & 0.01 & 2.201375e-08 & 2000.0 &    0.69 &    biased &  0.98 &      Scheme 2 \\
         L63 &  0.002 &  0.17 & 1.16 & 3.530047e-03 & 1833.0 &    2.36 &    biased &  0.98 &      Scheme 1 \\
         L63 &  5.985 &  0.60 & 0.80 & 8.493901e-08 & 2000.0 &    1.71 & quadratic &  0.98 &      Scheme 2 \\
      L9610d &  0.005 &  0.40 & 0.75 & 2.292198e-01 & 1550.0 &    0.44 &    biased &  0.98 &      Scheme 1 \\
      L9610d &  0.116 &  0.72 & 0.21 & 7.640822e-09 & 2000.0 &    1.57 &    biased &  0.98 &      Scheme 2 \\
       L96-5D &  0.004 &  0.87 & 0.14 & 1.972360e-08 & 1739.0 &    3.12 &    biased &  0.98 &      Scheme 1 \\
       L96-5D &  1.017 &  0.70 & 0.58 & 6.332524e-09 & 2000.0 &    1.73 &    biased &  0.98 &      Scheme 2 \\
    colpitts &  0.527 &  0.19 & 0.44 & 1.780703e-03 & 1897.0 &    1.19 &    biased &  0.98 &      Scheme 1 \\
    colpitts & 10.184 &  1.00 & 1.20 & 1.000000e-08 & 2000.0 &    2.00 &    linear &  0.98 &      Scheme 2 \\
     rossler &  0.712 &  0.67 & 0.68 & 1.533086e-02 & 1595.0 &    4.30 & quadratic &  0.98 &      Scheme 1 \\
     rossler &  2.204 &  0.47 & 0.50 & 2.101845e-09 & 2000.0 &    1.23 & quadratic &  0.98 &      Scheme 2 \\
    \bottomrule
    \end{tabular}
    \caption{Fig. \ref{fig: norm_results}}
    \label{tab:my_label}
\end{table}

\begin{table}[!htpb]
    \centering
    \begin{tabular}{lrrrrrrlrr}
    \toprule
     dataset & $\sigma$ & $\alpha$ &   $\rho_{SR}$ & $\beta$ & N & $\sigma_b$ & readout & $\rho_A$ & time step \\
    \midrule
    CL63 & 0.002 &  0.64 & 0.43 & 3.963368e-09 & 1500 &    1.63 &  linear &  0.98 &     0.005 \\
    CL63 & 0.004 &  0.96 & 0.28 & 8.664170e-07 & 1500 &    0.96 &  linear &  0.98 &     0.010 \\
    CL63 & 0.008 &  0.94 & 0.03 & 7.266341e-08 & 1500 &    1.85 &  linear &  0.98 &     0.050 \\
    CL63 & 0.004 &  0.61 & 1.55 & 1.322695e-02 & 1500 &    3.06 &  linear &  0.98 &     0.100 \\
     L63 & 0.020 &  1.00 & 1.10 & 3.671288e-08 & 1500 &    1.34 &  linear &  0.98 &     0.005 \\
     L63 & 0.034 &  0.92 & 0.69 & 1.294075e-08 & 1500 &    1.28 &  linear &  0.98 &     0.010 \\
     L63 & 0.056 &  0.79 & 0.18 & 5.401025e-10 & 1500 &    1.02 &  linear &  0.98 &     0.050 \\
     L63 & 0.040 &  0.79 & 0.05 & 1.436348e-07 & 1500 &    0.19 &  linear &  0.98 &     0.100 \\
      L9610d & 0.012 &  0.54 & 0.91 & 7.059347e-07 & 1500 &    1.82 &  linear &  0.98 &     0.005 \\
      L9610d & 0.020 &  0.51 & 0.87 & 4.387346e-08 & 1500 &    1.74 &  linear &  0.98 &     0.010 \\
      L9610d & 0.025 &  0.89 & 0.10 & 1.180616e-08 & 1500 &    2.37 &  linear &  0.98 &     0.050 \\
      L9610d & 0.003 &  0.61 & 0.75 & 1.237630e-02 & 1500 &    0.38 &  linear &  0.98 &     0.100 \\
       L96-5D & 0.014 &  0.95 & 1.64 & 1.604735e-07 & 1500 &    1.26 &  linear &  0.98 &     0.005 \\
       L96-5D & 0.066 &  0.90 & 1.31 & 1.672450e-09 & 1500 &    2.23 &  linear &  0.98 &     0.010 \\
       L96-5D & 0.025 &  0.67 & 0.07 & 1.654891e-10 & 1500 &    1.28 &  linear &  0.98 &     0.050 \\
       L96-5D & 0.100 &  0.69 & 0.03 & 1.903574e-08 & 1500 &    0.12 &  linear &  0.98 &     0.100 \\
    colpitts & 0.059 &  0.41 & 1.84 & 2.170770e-04 & 1500 &    1.69 &  linear &  0.98 &     0.005 \\
    colpitts & 0.100 &  0.83 & 2.00 & 5.228608e-05 & 1500 &    4.00 &  linear &  0.98 &     0.010 \\
    colpitts & 0.043 &  0.62 & 0.30 & 6.942212e-09 & 1500 &    0.15 &  linear &  0.98 &     0.050 \\
    colpitts & 0.100 &  0.98 & 1.95 & 1.000000e+00 & 1500 &    1.88 &  linear &  0.98 &     0.100 \\
     rossler & 0.050 &  0.66 & 0.23 & 8.252811e-09 & 1500 &    0.58 &  linear &  0.98 &     0.005 \\
     rossler & 0.030 &  1.00 & 1.52 & 2.060958e-08 & 1500 &    0.95 &  linear &  0.98 &     0.010 \\
     rossler & 0.035 &  0.43 & 2.00 & 2.999644e-10 & 1500 &    4.00 &  linear &  0.98 &     0.050 \\
     rossler & 0.081 &  0.60 & 1.48 & 1.025288e-10 & 1500 &    2.64 &  linear &  0.98 &     0.100 \\
    \bottomrule
    \end{tabular}
    \caption{Fig. \ref{fig: time step}}
    \label{tab: }
\end{table}

\begin{table}[!htpb]
\centering
\begin{tabular}{rrrrrrlr}
\toprule
$\sigma$ & $\alpha$ &   $\rho_{SR}$ &        $\beta$ &    $N$ & $\sigma_b$ & readout & $\rho_A$ \\
\midrule
0.001 &  0.71 & 0.93 & 9.988696e-01 & 1200 &    0.01 &  linear &  0.98 \\
0.002 &  0.32 & 0.78 & 5.411962e-02 & 2400 &    1.50 &  linear &  0.98 \\
0.002 &  0.21 & 0.80 & 9.887132e-01 & 3600 &    1.29 &  linear &  0.98 \\
0.001 &  0.90 & 1.50 & 1.000000e+00 & 4800 &    2.00 &  linear &  0.98 \\
0.001 &  1.00 & 0.07 & 1.758120e-07 & 6000 &    0.63 &  linear &  0.98 \\
\bottomrule
\end{tabular}
\caption{Fig. \ref{fig:why_scale_up}}
\label{tab:why_scale_up}
\end{table}

\begin{table}[!htpb]
\centering
\begin{tabular}{rrrrrrrrlr}
\toprule
$N_\text{output}$ & $N_\text{halo}$ & $\sigma$ & $\alpha$ & $\rho_{SR}$ &      $\beta$ &  $N$ & $\sigma_b$ & readout & $\rho_A$ \\
\midrule
                2 &               0 &    0.001 &     1.00 &        1.50 & 3.655278e-01 & 2000 &       1.11 &  linear &     0.98 \\
                2 &               2 &    0.027 &     0.67 &        1.18 & 1.221827e-08 & 2000 &       1.54 &  linear &     0.98 \\
                2 &               4 &    0.017 &     1.00 &        0.16 & 3.432201e-07 & 2000 &       0.75 &  linear &     0.98 \\
                2 &               6 &    0.005 &     0.94 &        0.46 & 6.443480e-07 & 2000 &       0.58 &  linear &     0.98 \\
                2 &               8 &    0.006 &     0.74 &        0.71 & 1.801521e-07 & 2000 &       1.23 &  linear &     0.98 \\
                4 &               0 &    0.004 &     0.45 &        0.39 & 2.475150e-01 & 2000 &       0.29 &  linear &     0.98 \\
                4 &               2 &    0.021 &     0.77 &        1.03 & 1.372422e-08 & 2000 &       1.27 &  linear &     0.98 \\
                4 &               4 &    0.003 &     0.99 &        0.64 & 4.054118e-07 & 2000 &       1.17 &  linear &     0.98 \\
                4 &               6 &    0.003 &     0.99 &        0.64 & 4.054118e-07 & 2000 &       1.17 &  linear &     0.98 \\
                4 &               8 &    0.003 &     0.99 &        0.64 & 4.054118e-07 & 2000 &       1.17 &  linear &     0.98 \\
                8 &               0 &    0.001 &     0.44 &        0.90 & 1.000000e+00 & 2000 &       1.74 &  linear &     0.98 \\
                8 &               2 &    0.003 &     0.99 &        0.64 & 4.054118e-07 & 2000 &       1.17 &  linear &     0.98 \\
                8 &               4 &    0.001 &     0.74 &        0.17 & 1.000000e-08 & 2000 &       0.72 &  linear &     0.98 \\
                8 &               6 &    0.003 &     0.99 &        0.64 & 4.054118e-07 & 2000 &       1.17 &  linear &     0.98 \\
                8 &               8 &    0.003 &     0.99 &        0.64 & 4.054118e-07 & 2000 &       1.17 &  linear &     0.98 \\
\bottomrule
\end{tabular}
\caption{Fig. \ref{fig:localrc}(a)}
\label{tab:localrc_a}
\end{table}

\begin{table}[!htpb]
\centering
\begin{tabular}{rrrrrrrrlr}
\toprule
$N_\text{output}$ & $N_\text{halo}$ & $\sigma$ & $\alpha$ & $\rho_{SR}$ &      $\beta$ &  $N$ & $\sigma_b$ & readout & $\rho_A$ \\
\midrule
                4 &               0 &    0.004 &     0.45 &        0.39 & 2.475150e-01 & 2000 &       0.29 &  linear &     0.98 \\
                4 &               2 &    0.021 &     0.77 &        1.03 & 1.372422e-08 & 2000 &       1.27 &  linear &     0.98 \\
                4 &               4 &    0.003 &     0.99 &        0.64 & 4.054118e-07 & 2000 &       1.17 &  linear &     0.98 \\
                4 &               6 &    0.003 &     0.99 &        0.64 & 4.054118e-07 & 2000 &       1.17 &  linear &     0.98 \\
                4 &               8 &    0.003 &     0.99 &        0.64 & 4.054118e-07 & 2000 &       1.17 &  linear &     0.98 \\
                4 &               0 &    0.004 &     0.45 &        0.39 & 2.475150e-01 & 1200 &       0.29 &  linear &     0.98 \\
                4 &               2 &    0.005 &     1.00 &        1.50 & 1.000000e-08 & 2400 &       2.00 &  linear &     0.98 \\
                4 &               4 &    0.005 &     0.36 &        0.48 & 1.000000e-08 & 3600 &       0.93 &  linear &     0.98 \\
                4 &               6 &    0.003 &     0.99 &        0.64 & 4.054118e-07 & 4800 &       1.17 &  linear &     0.98 \\
                4 &               8 &    0.003 &     0.99 &        0.64 & 4.054118e-07 & 6000 &       1.17 &  linear &     0.98 \\
\bottomrule
\end{tabular}
\caption{Fig. \ref{fig:localrc}(b)}
\label{tab:localrc_b}
\end{table}

\begin{table}[!htpb]
\centering
\begin{tabular}{lrrrrrrrrlr}
\toprule
                        Name & $N_\text{output}$ & $N_\text{halo}$ & $\sigma$ & $\alpha$ & $\rho_{SR}$ &      $\beta$ &  $N$ & $\sigma_b$ & readout & $\rho_A$ \\
\midrule
                    Local RC &                 2 &               2 &    0.011 &     0.49 &        0.85 & 1.155972e-08 &  720 &       1.11 &  linear &     0.98 \\
                   Single RC &                 40 &               0 &    0.001 &     1.00 &        0.07 & 1.758120e-07 & 6000 &       0.63 &  linear &     0.98 \\
Model 3 from \cite{penny2021} &                 2 &               4 &    0.052 &     0.41 &        0.34 & 3.611497e-06 & 6000 &       0.00 &  linear &     0.99 \\
\bottomrule
\end{tabular}
\caption{Fig. \ref{fig:performance_showdown}}
\label{tab:performance_showdown}
\end{table}

\newpage
\section{Minimal RC Code Julia}
\begin{jllisting}
using SparseArrays
using Distributions
using LinearAlgebra
using Random
using Arpack

struct esn 
    D::Int # Input data dimension
    N::Int # Reservoir dimension
    SR::Float64 # Spectral Radius of A
    ρA::Float64 # Density of A
    α::Float64 # Leak rate
    σ::Float64 # Input strength 
    σb::Float64 # Bias
    β::Float64 # Tikhonov regularization
    A::SparseMatrixCSC{Float64, Int64} # Adjacency Matrix
    Win::Array{Float64,2} # Input matrix
    Wout::Array{Float64,2} # Output matrix
    rng::MersenneTwister # random seed
    function esn(D, N, SR, ρA, α, σ;
                 random_state=111111, σb=0, β=1e-12)
        rng = MersenneTwister(random_state)
        A = get_connection_matrix(N, ρA, SR, rng)
        Win = rand(rng, Uniform(-σ, σ), N, D)
        Wout = Array{Float64}(undef, (D, N)) 
        new(D, N, SR, ρA, α, σ, σb, β, A, Win, Wout, rng)
    end
end

function get_connection_matrix(N, ρA, SR, rng)
    rf(rng, N) = rand(rng, Uniform(-1.0, 1.0), N)
    A = sprand(rng, N, N, ρA, rf)
    eig, _ = eigs(A, nev=1)
    maxeig = abs(eig[1])
    A = A.*(SR/maxeig)
    return A
end

function train_RC(esn::esn, u, spinup)
    r = generate(esn, u)
    compute_Wout(esn, r[:, 1+spinup:end], u[:, 1+spinup:end])
end

function forecast_RC(esn::esn, nsteps; uspin=nothing, r0 = nothing)
    if !isnothing(uspin)
        rspin = generate(esn, uspin)
        r0 = rspin[:, end]
    end
    @assert !isnothing(r0)

    rfc = Array{Float64}(undef, (esn.N, nsteps))
    rfc[:, 1] = r0
    
    for t in 1:nsteps-1
        auto_esn(rfc[:, t+1], esn, rfc[:, t])
    end
    return esn.Wout*rfc
end

function auto_esn(rtp1, esn::esn, rt)
    rtp1[:] .= esn.α.*tanh.(esn.A*rt .+ esn.Win*esn.Wout*rt .+ esn.σb).+(1 .- esn.α).*rt
end

function driven_esn(rtp1, esn::esn, rt, ut)
    return rtp1[:] .= esn.α.*tanh.(esn.A*rt .+ esn.Win*ut .+ esn.σb).+(1 .- esn.α).*rt
end

function generate(esn::esn, u)
    T = size(u)[2]
    r = zeros(esn.N, T)
    for t in 2:T
        driven_esn(r[:, t], esn, r[:, t-1], u[:, t-1])
    end
    return r
end

function compute_Wout(esn::esn, r, u)
        esn.Wout[:, :] .= ((r*r'+esn.β*I) \ (r*u'))'
end

\end{jllisting}

\end{document}

%% file: defs.tex
\newcommand{\eg}{{\it e.g.}}
\newcommand{\ie}{{\it i.e.}}

\newcommand{\Win}{\mathbf{W_{\rm in}}}
\newcommand{\Wout}{\mathbf{W_{\rm out}}}

\def\bphi{\bm{\phi}}
\def\J{\vb J}
\def\r{\vb r}
\def\x{\vb x}
\def\A{\vb A}
\def\B{\vb B}
\def\u{\vb u}
\def\W{\vb W}
\def\I{\mathbb{I}}
\def\M{\mathcal{M}}
\def\DF{\mbox{{\bf DF}}}
\def\diag{\rm{diag}}

\definecolor{gblue}{rgb}{0.0,0.7,0.6}